\documentclass[preprint]{elsarticle}
\usepackage{geometry}  
\usepackage{lineno,hyperref}
\usepackage{graphicx}
\usepackage{multirow}
\usepackage{graphicx}
\usepackage{caption}
\usepackage{xcolor}
\usepackage{tabularx}
\usepackage{adjustbox}
\usepackage{cite}
\usepackage{longtable}
\usepackage{subfigure}
\usepackage{float}
\modulolinenumbers[5]

\usepackage[ruled,vlined]{algorithm2e}
\SetAlFnt{\scriptsize}

\journal{WIREs Data Mining and Knowledge Discovery}

\biboptions{authoryear}

\begin{document}

\begin{frontmatter}

\title{A survey of textual cyber abuse detection using cutting-edge language models and large language models}

\author[inst1]{J. Angel Diaz-Garcia orcid: 0000-0002-9263-1402 }
\author[inst2]{Joao Paulo Carvalho orcid: 0000-0003-0005-8299}

\affiliation[inst1]{organization={Department of Computer Science and A.I, University of Granada},    
            addressline={C. Periodista Daniel Saucedo Aranda, s/n}, 
            city={Granada},
            postcode={18014}, 
            state={Granada},
            country={Spain }, email={joseangeldiazg@ugr.es}}

\affiliation[inst2]{organization={INESC-ID, Instituto Superior Técnico, Universidade de Lisboa},
            addressline={Av. Rovisco Pais 1}, 
            city={Lisboa},
            postcode={1049-001}, 
            state={Lisboa},
            country={Portugal },  email={joao.carvalho@inesc-id.pt}}

\begin{abstract}
	
The success of social media platforms has facilitated the emergence of various forms of online abuse within digital communities. This abuse manifests in multiple ways, including hate speech, cyberbullying, emotional abuse, grooming, and sexting. In this paper, we present a comprehensive analysis of the different forms of abuse prevalent in social media, with a particular focus on how emerging technologies, such as Language Models (LMs) and Large Language Models (LLMs), are reshaping both the detection and generation of abusive content within these networks. We delve into the mechanisms through which social media abuse is perpetuated, exploring the psychological and social impact. Additionally, we examine the dual role of advanced language models-highlighting their potential to enhance automated detection systems for abusive behavior while also acknowledging their capacity to generate harmful content. This paper aims to contribute to the ongoing discourse on online safety and ethics, offering insights into the evolving landscape of cyberabuse and the technological innovations that both mitigate and exacerbate it.
\end{abstract}


\end{frontmatter}


\section*{Declarations}

\begin{itemize}
\item \textbf{Funding information:} This research was funded by the European Union, the Spanish Ministry of Science, Innovation, and Universities and the University Of Granada.
\item \textbf{Conflict of Interest:} The authors declare that they have no conflicts of interest relevant to the content of this article.
\item \textbf{Consent for Publication:} All authors have reviewed and approved the manuscript and consent to its publication.
\item \textbf{Data availability:} Data sharing is not applicable to this article as no new data were created or analyzed in this study
\end{itemize}

\section{Introduction}
\label{sec:introduction}

In the digital age, social networks and online platforms have become integral to everyday life, facilitating communication, information sharing, and social interaction. However, this digital connectivity has also given rise to various forms of cyber abuse, affecting millions of individuals globally. The pervasive nature of cyber abuse, including hate speech, cyberbullying, emotional abuse, doxxing, trolling, impersonation, and shaming, poses significant risks to mental health and well-being causing psychosocial problems \citep{kwan2020cyberbullying} as anxiety or depression \citep{fisher2016peer}.

Traditional methods of addressing cyberabuse have often relied on manual reporting systems and heuristic-based detection \citep{vandebosch2009cyberbullying}. However, recent advancements in artificial intelligence (AI), particularly through LLMs, offer new avenues for understanding and combating these issues. LLMs, such as GPT \citep{radford2018improving}, BERT \citep{Devlin2018}, and other state-of-the-art models \citep{raffel2020exploring,liu2019roberta}, have demonstrated remarkable capabilities in natural language processing, providing powerful tools for analyzing and detecting harmful content.

This paper aims to shed light on the diverse forms of cyberabuse through the lens of these cutting-edge AI techniques. By leveraging LLMs and advanced data analysis methods, we seek to explore how these technologies can enhance the detection, classification, and mitigation of various types of cyberabuse. The importance of this research is underscored by the profound impact cyberabuse has on individuals and communities, even leading to self-harm and suicidal thoughts \citep{hamm2015prevalence,kim2019sex,kowalski2014bullying}. Despite the progress made in addressing more prominent issues like hate speech and cyberbullying, there remains a significant gap in understanding and tackling other forms of abuse.

We provide a comprehensive overview of how these models can be employed to address both well-researched and less-explored types of cyberabuse. By doing so, we highlight the potential of AI to offer more nuanced and effective solutions to combat cyberabuse, ultimately contributing to a safer and more supportive online environment.

In recent years, several surveys have explored hate speech detection \citep{fortuna2018survey}; however, there is a noticeable gap in covering the latest advancements, particularly the emergence of cutting-edge models like LLMs that have been released in the past year. Therefore, our primary goal is to update the existing literature with these recent developments. Additionally, since most studies in this domain approach the problem as a classification task, we aim to conduct an in-depth analysis of the evaluation processes, especially considering the challenges posed by the unbalanced nature of cyber abuse detection. Moreover, existing reviews predominantly concentrate on cyberbullying \citep{salawu2017approaches} or hate speech \citep{paz2020hate,schmidt2017survey}, leaving other forms of cyber abuse significantly underrepresented. With this in mind, the main contributions of our paper are as follows:

\begin{itemize}
    \item We present the most up-to-date analysis of textual cyber abuse, with a particular emphasis on cutting-edge techniques that leverage  LMs and LLMs, identifying current gaps and suggesting avenues for future research and improvement. 
    \item To the best of our knowledge, we conduct the first comprehensive survey that addresses a broader spectrum of cyber abuse forms, including doxing, trolling, and impersonation, which have been largely overlooked in previous reviews. 
    \item We perform an in-depth analysis of the evaluation metrics used in cyber abuse detection, paying special attention to the challenges posed by class imbalance, a common issue in this domain.
\end{itemize}

To address these questions, the paper is organized as follows: Section \ref{sec:forms_of_textual_cyberabuse} examines the various forms of textual cyber abuse encountered in online social interactions. Section \ref{sec:impact_of_language_models} analyzes the impact of LMs and LLMs in two main aspects: their role in detecting cyber abuse and hate speech, as well as their potential in generating automated instances of such content. Section \ref{sec:other} delves into how artificial intelligence is advancing the detection and analysis of other, less-explored forms of cyber abuse. Finally, we present a comprehensive discussion in Section \ref{sec:discussion} and draw our conclusions in Section \ref{sec:conclusion}.

\section{Forms of Textual Cyberabuse}
\label{sec:forms_of_textual_cyberabuse}

In this section, we explore the various forms of textual cyberabuse prevalent in social networks. Each subsection delves into the definition, characteristics, and impact of a specific type of abuse. 

\begin{figure}[h]
  \centering
  \includegraphics[width=0.7\textwidth]{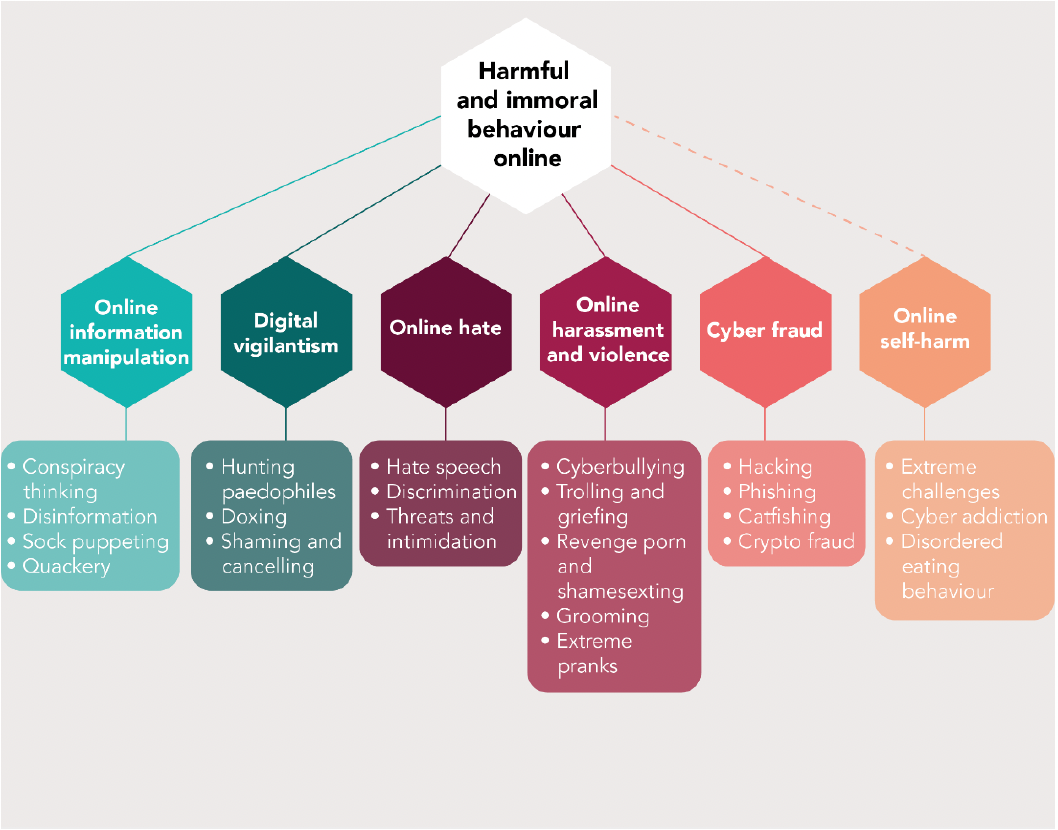}
  \caption{Harmful and immoral behaviour online}
  \label{fig:harnful}
\end{figure}

In \citep{van2022harmful}, the authors identify a range of harmful and immoral uses of social media, categorizing these into six primary types, which can be further analyzed into 22 distinct tasks. The full categorization is depicted in Figure \ref{fig:harnful}, reproduced from \citep{van2022harmful}.

It is important to note that the advent of generative AI has accelerated certain forms of cyberabuse, such as the creation of fake sexual content, the generation of misleading images, and the widespread dissemination of both authentic and fabricated content across social networks. However, these forms of abuse fall outside the scope of this paper, as we focus specifically on textual content. Therefore, our analysis is centered on various types of text-based cyberabuse.

To align with existing research in the field, we have selected the following subgenres of cyberabuse in social media for our analysis: hate speech, cyberbullying, doxing, shaming, cancel culture, and trolling as the primary components. Additionally, we include other forms of cyberabuse such as impersonation \citep{gharawi2021social}, and emotional abuse \citep{stephenson2018psychological}. In the following subsections, we introduce each of these subgenres in detail.

\subsection{Hate Speech}
\label{subsec:hate_speech}

Hate speech involves language that demeans, intimidates, or incites violence against individuals or groups based on attributes such as race, religion, ethnicity, gender, sexual orientation, or other characteristics. This form of expression poses severe threats to individual dignity, democratic values, social stability, and peace. Despite its profound impact, there remains a general lack of awareness regarding the extensive harm that hate speech can inflict across various aspects of life.

Hate speech on social media is a pressing social issue. Research indicates that exposure to such speech can lead to a range of mental health problems, particularly among young adults, as well as psychosocial issues \citep{kwan2020cyberbullying}. Consequences may include anxiety \citep{fisher2016peer}, depression \citep{fisher2016peer,hamm2015prevalence,wachs2022online}, and post-traumatic stress disorder \citep{cassiani2022factors,liu2020longitudinal,wypych2022psychological}.

Detecting hate speech requires analyzing text for derogatory terms, slurs, and contextual indicators of malicious intent. NLP techniques, and more recently LMs and LLMs, are crucial tools in this effort. These technologies can automatically capture the specific nuances of hate speech in social networks with high accuracy. This survey focuses on recent advancements in hate speech detection, particularly the integration of new language models to enhance detection capabilities.

\subsection{Cyberbullying}
\label{subsec:cyberbullying}

Cyberbullying affects both children and adolescents and is widely recognized as intentional, violent, cruel, and repetitive behavior directed at peers. This pervasive issue is identified by the US Centers for Disease Control and Prevention (CDC) as a significant public health threat \citep{centers2009technology}. The emotional and psychological damage caused by cyberbullying can be severe, with extreme cases potentially leading to suicide \citep{bauman2013associations}.

According to \citep{mcloughlin2009texting}, cyberbullying can be categorized into several types, including flaming, harassment, denigration, impersonation, outing, boycott, and cyberstalking. Particularly in more severe forms such as flaming and harassment, cyberbullying often manifests through texts, comments, or messages exchanged between the bully and the victim. These communications typically involve insults and attacks related to race, sexuality, ethnicity, or social status. Such messages represent user-generated content in an unstructured format, which can be analyzed automatically using AI systems based on NLP and LLMs to identify and address hate speech.

In the current digital era, cyberbullying has evolved to encompass a wide range of multimodal categories, including the creation of sexual deep fakes targeting individuals \citep{rini2022deepfakes,rousay2023sexual,dunn2024legal}. While this is a significant area of concern, our focus remains on the textual aspects of cyberbullying, specifically analyzing comments and interactions on social networks and online platforms.

\subsection{Emotional and Psychological Abuse}
\label{subsec:emotional_psychological_abuse}

This form of abuse involves manipulative or controlling language aimed at undermining an individual's emotional well-being. Examples include gaslighting, persistent criticism, and verbal assaults. While emotional and psychological abuse often overlaps with cyberbullying, especially since all forms of cyberbullying inherently involve emotional harm, it is important to distinguish between the two. Cyberbullying is more commonly associated with early life stages, such as during school or adolescence, whereas psychological abuse can occur at any stage of life, including in relationships with partners in adulthood \citep{moral2022emotional}. So in our survey, we aim to establish a more fine-grained distinction by categorizing and summarizing various forms of cyberbullying and their intersections with artificial intelligence, with a particular focus on LLMs. Regarding psychological abuse, our inclusion criteria specifically exclude cases of psychological abuse occurring during childhood or other life stages unrelated to online interactions \citep{WOS:000968623100040}, as our focus is solely on cyber abuse, and the real of these kind of applications are more related to medicine area \citep{WOS:000819681100017} or psychology \citep{WOS:001245042700001}.

\subsection{Doxing}
\label{subsec:doxxing}

Doxing is the act of publicly disclosing private or personally identifiable information about an individual without their consent \citep{karimi2022automated}. This can manifest in various forms, such as posts or messages that share sensitive data, including addresses, phone numbers, and other confidential details. Often, doxing is akin to a data leak, and it is typically executed with malicious intent, aiming to harm, intimidate, or harass the victim. The harmful effects of doxing on social networks are significant and include:

\begin{itemize}
    \item \textbf{Privacy Violations}: Doxing infringes on an individual’s right to privacy by exposing their personal information to the public, which can lead to unwanted attention and potential harassment.
    \item \textbf{Emotional and Psychological Impact}: Victims of doxing may experience severe emotional distress, including anxiety, depression, and a heightened sense of vulnerability due to the public exposure of their private information.
    \item \textbf{Professional Consequences}: Doxing can have detrimental effects on a person's career, damaging professional relationships, leading to job loss, and causing reputational harm.
\end{itemize}

In the literature, it is essential to distinguish between doxing experienced by individuals and that which affects organizations \citep{herrera2022survey}. This paper focuses exclusively on the impact of doxing on individuals. Finally, it is necessary to mention that data leakage in the realm of machine learning can be related to the unintentional inclusion of information from the test set in the training process, this topic falls out of our analysis.

\subsection{Trolling}
\label{subsec:trolling}

According to Bishop \citep{bishop2013art}, trolling is defined as ``the activity of posting messages via a communications network that are intended to be provocative, offensive, or menacing.'' This definition highlights that trolling involves posting inflammatory or irrelevant messages with the intent to provoke or disrupt online conversations.

In the literature, various approaches categorize trolling not only by the content of the messages but also by the underlying intentions. For example, De et al. \citep{de2018modeling} identify four key aspects of trolling:

\begin{itemize}
    \item \textbf{Intention (I):} The author's underlying purpose or motive behind the trolling activity. \item \textbf{Intention Disclosure (D):} This aspect evaluates whether the author is concealing their true (malicious) intentions from the audience. \item \textbf{Intention Interpretation (R):} This reflects how responders perceive and interpret the troll's intentions. \item \textbf{Response Strategy (B):} This describes the responder's reaction to the trolling attempt, which may also be a form of trolling. 
\end{itemize}

Trolling is a significant issue that requires attention, as highlighted by De et al. \citep{de2018modeling}. Their study found that in 26.9\% of trolling incidents, the victims engage with the trolling attempt, often experiencing some form of emotional distress.

In our survey, we concentrate on the core aspects of trolling within social network conversations, specifically examining how trolling manifests through interactions in textual dialogues. We exclude from our analysis newer forms of trolling, such as those involving memes, as they represent a different avenue of exploration that falls outside the direct scope of conversational interactions \citep{hossain2022identification,suryawanshi2023trollswithopinion}.

\subsection{Impersonation}
\label{subsec:impersonation}

Impersonation involves creating fake profiles or hacking into existing accounts to deceive others. Typically, impersonators target brands or well-known individuals, aiming to extract some form of value through their deception \citep{zarei2020impersonation}. 

Numerous studies have explored the issue of impersonation, with a particular focus on understanding its impact and vulnerabilities. The findings from Yu et al. \citep{yu2023vulnerability} on the susceptibility of older adults to government impersonation scams highlight alarming trends that call for urgent, targeted interventions. Approximately 16\% of participants engaged in conversations with an agent impersonating a government representative without displaying skepticism, indicating a high risk of victimization within this demographic. Even more concerning, 12\% of participants willingly shared personal information, and nearly 5\% provided the last four digits of their Social Security number. While impersonation has been widely recognized and studied as a growing problem, efforts to detect and prevent these scams remain inadequate and underdeveloped, leaving vulnerable populations at continued risk.

Impersonation often relies on the use of altered images, deep fakes, or even biometric information of the scammed individuals \citep{liu2024eap}. Also we found a stream regarding the impersonations and infiltration between networks, as a kind of network attack \citep{shukla2023detecting}.  In this survey, we have excluded these types of impersonation detection methods, as they fall within the domain of image analysis. Our focus is on text-based detection based in textual indicators or features as sudden changes in language style or tone, and inconsistencies in personal information.


%


\subsection{Shaming and Cancel Culture}
\label{subsec:shaming_cancel_culture}

Shaming and cancel culture involve public criticism or ostracism of individuals based on their actions or statements, sometimes leading to online harassment. Public Shaming refers to criticizing or condemning individuals or groups in a public forum, often through social media posts, comments, or viral campaigns. This practice aims to invoke feelings of guilt or shame and can escalate quickly, resulting in widespread condemnation from the online community \citep{basak2019online}. Cancel Culture involves withdrawing support from individuals, organizations, or brands after they have made statements or engaged in actions deemed objectionable or offensive. Harmful Implications of Shaming and Cancel Culture:

\begin{itemize}
    \item \textbf{Emotional and psychological distress}: Victims may experience significant emotional and psychological impacts, such as anxiety, depression, and a sense of isolation \citep{muir2023examining}.
    
    \item \textbf{Effect on free expression}: The fear of public shaming or being "canceled" can discourage individuals from expressing their opinions or participating in discussions on important issues \citep{koivukari2021online}.
    
    \item \textbf{Normalization of aggressive behavior}: These practices can foster a hostile online environment, where aggressive behavior becomes normalized and further encouraged \citep{beres2021don}.
\end{itemize}

Although there is significant research addressing issues related to social networks such as Twitter, our survey specifically focuses on recent advancements involving language models or neural models published in academic journals. Consequently, some notable studies, such as those by Basak et al. \citep{basak2019online}, Surani et al. \citep{surani2021comparative}, and Nalawade et al. \citep{nalawade2021survey}, are not included in our survey.

\section{Methodology}

In this paper, we applied the PRISMA methodology \citep{moher2010preferred} to conduct a thorough and systematic review aligned with our research objectives. The PRISMA framework comprises four key stages: formulating research questions, developing a search strategy, defining eligibility criteria, and selecting relevant studies.

\subsection{Research questions}

\begin{itemize}
    \item  \textbf{RQ1}: How representative are LMs and LLMS in detecting textual cyber abuse compared to traditional methods?
    \item \textbf{RQ2}:Considering that cyber abuse frequently involves imbalanced classification challenges, are researchers adequately addressing this issue in their evaluation metrics and pre-processing methods?
     \item \textbf{RQ3}: Based on the reviewed literature, what emerging trends and future directions are shaping the development of cyber abuse detection?
    \item \textbf{RQ4}: Which types of cyber abuse detection are most significantly influenced by LLMs?
    \item \textbf{RQ5}: Are there specific forms of cyber abuse that remain underexplored and lack adequate detection solutions?
\end{itemize}

\subsection{Search strategy}

The vast majority of research in the field of textual cyber abuse detection focuses on hate speech and cyberbullying. To capture the most recent advancements, our search strategy is twofold. For hate speech, trolling and cyberbullying, we have narrowed the results to papers that address these issues using cutting-edge techniques based on LLMs. In contrast, for other forms of abuse-where research is less prominent-we have broadened our queries to include additional analysis-related terms. To ensure robust and comprehensive results, we focused our analysis on the Web of Science (WOS) database. This database exclusively indexes peer-reviewed papers and encompasses other databases, such as IEEE Xplore and SpringerLink, allowing us to cover multiple indexed sources while benefiting from WOS's advanced search capabilities. For each topic, we used the following research queries.

 \begin{itemize}
    \item \textbf{Hate Speech:} \\
    \texttt{TI=("hate speech" OR "hate content" OR "hate messages") AND AB=("transformers" OR "GPT" OR "BERT" OR "T5" OR "LLMs") AND TS=("detection" OR "classification" OR "analysis")}

    \item \textbf{Cyberbullying:} \\
    \texttt{TI=("cyberbullying" OR "online harassment" OR "cyber harassment”) AND AB=("transformers" OR "GPT" OR "BERT" OR "T5" OR "LLMs") AND TS=("detection" OR "classification" OR "prediction" OR "analysis")}

    \item \textbf{Emotional and Psychological Abuse:} \\
    \texttt{(TI=("emotional abuse" OR "psychological abuse" OR "verbal abuse" OR "emotional manipulation" OR "mental abuse" OR "emotional distress" OR "psychological distress"  OR "emotional harm" OR "psychological harm" OR "emotional violence" OR "psychological violence" OR "verbal harassment" )) AND AB=("artificial intelligence" OR "natural language processing" OR "deep learning" OR transformers OR GPT OR BERT OR LLAMA OR 'neural networks')}

    \item \textbf{Doxing:} \\
    \texttt{(TI=(("doxing" OR "doxxing" OR "doxing attack" OR "doxxing attack" OR "personal information exposure" OR "personal data exposure" OR "data leakage" OR "identity exposure" OR "identity disclosure" OR "public exposure" OR "private information leak" OR "information leak" OR "confidential information disclosure" OR "online doxing" OR "online doxxing" OR "Privacy-Disclosure" OR "PrivacyLeak" OR "private information")) OR AB=(doxing)) AND TS=("artificial intelligence" OR "machine learning" OR "natural language processing" OR "data analysis" OR "deep learning” OR transformers OR GPT OR BERT OR LLAMA OR 'neural networks') AND TS=("detection" OR "prevention" OR "analysis")}

    \item \textbf{Trolling:} \\
    \texttt{(TI=(trolling OR "online trolling" OR "internet trolling") ) AND TS=("artificial intelligence" OR "machine learning" OR "natural language processing" OR "data analysis" OR "deep learning” OR transformers OR GPT OR BERT OR LLAMA OR 'neural networks')}

    \item \textbf{Impersonation:} \\
    \texttt{(TI=("impersonation" OR "identity theft" OR "fake profiles” OR   "catfishing" OR  "fake social media accounts ") ) AND TS=("artificial intelligence" OR "machine learning" OR "natural language processing" OR "data analysis" OR "deep learning” OR transformers OR GPT OR BERT OR LLAMA OR 'neural networks')}


    \item \textbf{Shaming and Cancel Culture:} \\
    \texttt{TI=("shaming" OR "cancel culture" OR "public shaming" OR "online shaming" OR "social shaming" OR "digital shaming" OR "shaming campaigns" OR "social ostracism" OR "public humiliation" OR "online humiliation" OR "reputation damage" OR "character assassination" OR "social backlash" OR "public backlash" OR "mob justice" OR "cyber shaming" OR "call-out culture" OR "name and shame" OR "internet shaming" OR "public reprimand")  AND AB=("artificial intelligence" OR "machine learning" OR "natural language processing" OR "data analysis" OR "deep learning" OR transformers OR GPT OR BERT OR LLAMA OR 'neural networks’) AND TS=("artificial intelligence" OR "machine learning" OR "natural language processing" OR "data analysis" OR "deep learning") AND TS=("detection" OR "prevention" OR "analysis")}
\end{itemize}

\subsection{Eligibility criteria}

Regarding the eligibility criteria, we have focused on journal papers published in the last three years (January 2022- September 2024), as this period aligns with the significant advancements and widespread adoption of LLMs. We selected papers that primarily involve text-based datasets, ensuring relevance to our focus on LLMs. While some papers may also address social network analysis through graph theory or image detection, they were only included if they incorporated a textual analysis component. For instance, a paper primarily centered on graph theory but also analyzing tweet content was considered. Additionally, we excluded survey-based studies focusing on psychological assessments through participant responses, as our interest lies in the direct analysis of textual data rather than self-reported experiences. Table \ref{tab:inclusion_results} provides a summary of the inclusion criteria and the final outcomes of the selected papers. 

\begin{table}[htb]
\centering
\renewcommand{\arraystretch}{1.5} 
\begin{adjustbox}{max width=\textwidth}
\begin{tabular}{p{4cm}p{2.5cm}p{3cm}p{3cm}p{6cm}p{2.5cm}}
\hline
\textbf{Category} & \textbf{Inclusion Years} & \textbf{Document Type} & \textbf{Raw Results} & \textbf{Inclusion Criteria} & \textbf{Final Results} \\ 

\hline
Hate Speech & 2022-2024 & Article & 41 & Text and online related, accessible & 37 \\ 
Cyberbullying & 2022-2024 & Article & 17 & Text and online related, accessible & 15 \\ 
Shaming and Cancel Culture & 2022-2024 & Article & 1 & Text based and online related, accessible & 1 \\ 
Doxing & 2022-2024 & Article & 9 & Text-based and individuals' information related, private information related, accessible & 3 \\ 
Trolling & 2022-2024 & Article & 13 & Text based in online environments. Not meme-based, accessible & 8 \\ 
Impersonation & 2022-2024 & Article & 6 & Text based and online related, accessible & 1 \\ 
Emotional and Psychological abuse & 2022-2024 & Article & 11 & Text based with possible applications in cyber abuse detection, accessible & 3 \\ 
\hline
\end{tabular}
\end{adjustbox}
\caption{Inclusion Criteria and Results Summary}
\label{tab:inclusion_results}
\end{table}

\section{Impact of Language Models on Hate Speech and cyberbullying detection}
\label{sec:impact_of_language_models}

In this section, we conduct an analysis focusing on the adoption of different LMs and LLMs for hate speech and cyberbullying detection. To enhance clarity, the section is divided into two parts: hate speech detection  (Section \ref{sec:hate-detection}) and cyberbullying detection (Section \ref{sec:cyber-detection}). Acknowledging the rise of generative AI being used for harmful purposes, we have also included a section that explores the generation of abusive content (Section \ref{sec:generation}).

\subsection{Hate Speech detection}
\label{sec:hate-detection}

The complexities of hate speech often necessitate the use of a combination of techniques and models to effectively address each specific task. In this context, Khan et al. \citep{khan2022bichat} proposed a system that utilizes BERT for embeddings, a CNN to capture spatial features of the text, and a BiLSTM to model sequential dependencies. The model also incorporates hierarchical attention, which assigns different weights to features extracted from both the CNN and BiLSTM outputs. This hierarchical attention mechanism is a key contribution, featuring two levels of attention: a high-level focus on the CNN outputs and a low-level focus on the BiLSTM outputs. 

Gupta et al. \citep{gupta2021bert} highlighted that the main contribution of their paper was the application of BERT to the problem of hate speech detection. However, we found the most compelling aspect to be their comprehensive comparative analysis across four different datasets, with particular attention to fine-tuning and parameter adjustments. The authors achieved good results and provided insightful interpretations for each dataset.


Hate speech frequently targets specific groups or individuals based on characteristics such as race, gender, or sexual orientation. To address this issue, some studies have adopted multi-task, multi-aspect, or dual-objective approaches, with ensemble methods often outperforming state-of-the-art models in this task. Mazari et al. \citep{mazari2024bert} introduced novel architectures that combine the pre-trained BERT model with deep learning techniques, specifically leveraging ensemble learning strategies that incorporate Bi-LSTMs and Bi-GRUs. Their approach effectively tackles multi-aspect hate speech, covering various forms such as identity-based hate and threats. In \citep{yun2023bert}, Yun et al. tackle both hate speech detection and gender bias, which refers to the unfair or prejudiced treatment of individuals based on their gender. The authors proposed a BERT-based ensemble model that integrates various pre-trained models, including Soongsil-BERT, KcELECTRA, BERT, and RoBERTa, to enhance detection accuracy. The research emphasizes the effectiveness of combining multiple pre-trained models using a weighting scheme that accounts for label distribution, improving performance on imbalanced datasets. To further address class imbalance, they applied data oversampling techniques, allowing the model to learn more effectively from underrepresented classes. Karayığit et al. \citep{karayiugit2022homophobic} made a similar contribution by focusing on homophobic and hate speech detection in the Turkish language. Their work stands out as one of the first contributions in this area, utilizing an M-BERT model for this task. In \citep{malik2024hate}, the author also delves into the multifaceted nature of hate speech by not only developing a language model-based classifier for hate speech detection but also conducting community detection to gain deeper insights into the communities most affected by hate speech. The study particularly focuses on religious, political, and ethnic communities, offering a comprehensive approach to understanding both the prevalence of hate speech and the groups that are disproportionately targeted.

In \citep{su2023ssl}, Su et al. proposed a semi-supervised system for hate speech detection that learns from a limited set of annotated examples while leveraging a larger pool of unannotated data. This approach is particularly beneficial in scenarios where labeled data is scarce. The model is built using a Generative Adversarial Network (GAN) in conjunction with RoBERTa as the encoder. The GAN consists of two primary components: a generator and a discriminator. The generator creates synthetic data that closely mimics the real data, while the discriminator evaluates the authenticity of the generated data in comparison to real data. This architecture enables the model to improve its ability to detect hate speech. The results show that the proposed model achieves significant performance gains over the baseline RoBERTa model, particularly in terms of accuracy and Macro-F1, highlighting its effectiveness in semi-supervised learning scenarios. In a similar study, \citep{putra2024semi} tackled hate speech detection using a semi-supervised approach by employing BERT as an encoder alongside a CNN architecture composed of three parallel branches to process the input embeddings. The authors introduced a method of generating pseudolabels for unlabeled data, where a model trained on the labeled data predicts labels for unlabeled instances, thereby expanding the training set. A notable enhancement in their model is the inclusion of a shared BERT layer that enables the model to learn from multiple datasets concurrently. This shared layer works alongside private layers dedicated to individual tasks; the shared layer captures generalized features relevant across tasks, while the private layers focus on task-specific nuances. The datasets used for training included Hatebase, Supremacist, Cybertroll, TRAC, and TRAC 2020. The BERT-3CNN model demonstrated substantial performance improvements, achieving an F1 score enhancement of 18\% over state-of-the-art models.


One significant challenge in hate speech and other forms of cyber abuse detection is the lack of sufficient labeled examples to train robust models. Addressing this issue, while also aiming to improve the explainability of typically opaque models, \citep{ansari2024data} investigates the impact of three distinct data augmentation techniques on two publicly available datasets: one containing comments from Facebook and YouTube, and another composed of tweets from Twitter, all labeled for hatefulness or neutrality. The augmentation techniques applied include Easy Data Augmentation, BERT-based paraphrasing, and Back Translation. EDA involves simple text manipulation methods to create additional training samples. BERT-based augmentation leverages the model to generate paraphrases by predicting masked words or producing semantically equivalent sentences. Back Translation translates text into another language and back into the original to produce syntactically varied but semantically similar sentences. For classification, the authors utilized Convolutional Neural Networks (CNNs) and Long Short-Term Memory (LSTM) networks. This approach not only addresses the data scarcity issue but also reduces the risk of overfitting, enhancing the models' generalization capabilities and improving their performance in real-world hate speech detection scenarios. Wullach et al. \citep{wullach2022character} proposed an intriguing approach that leverages HyperNetworks, a type of neural network architecture designed to generate weights for another neural network, enabling efficient weight sharing across layers and operating at the character level. This character-level focus allows the model to effectively manage morphological variations and the noisy text commonly found in social media. Additionally, using GPT-2, the authors generated a large-scale corpus of synthetic hate and non-hate sequences to augment existing datasets, effectively addressing challenges related to limited labeled data and class imbalance.

One of the most comprehensive datasets for hate speech detection is presented by Toliyat et al. \citep{toliyat2022asian}, comprising over 10 million tweets related to COVID-19 and hate speech targeting Asian communities. The authors collected this dataset by crawling Twitter using specific keywords. To establish a reliable ground truth, they manually annotated 3,000 hate speech examples from the dataset proposed by He et al. \citep{he2021racism}. This manually labeled subset was then leveraged to label the entire 10 million tweets. The study also conducts an extensive evaluation of various machine learning approaches, including deep learning, transformer models, and traditional machine learning techniques, across different configurations of balanced and imbalanced datasets.

To address the scarcity of hate speech research in languages other than English, which have over 60\% of efforts in this field \citep{pikuliak2021cross}, Firmino et al. \citep{firmino2024improving} proposed the use of cross-lingual methodologies. The authors first gathered data from English and Italian, leveraging these larger datasets to train language models such as BERT. In a subsequent step, they fine-tuned the model on the target language, Portuguese, using smaller datasets. The results demonstrated the effectiveness of this cross-lingual approach, achieving an F1-measure of 92\% on the OffComBr-2 corpus. The study's findings highlight that incorporating Latin-based languages, like Italian, significantly improves hate speech detection in Portuguese, showcasing the potential of transfer learning in enhancing performance for low-resource languages.


In \citep{zhou2023automated}, Zhou et al. introduced a manually annotated dataset comprising posts from HackForums, Stormfront, and Incels.co, focusing on the prevalence and nature of hate speech across these platforms. The study revealed that while underground forums like HackForums exhibited lower levels of hate speech, extremist platforms had significantly higher volumes, driven by specific ideologies. To address the challenge of hate speech detection, the authors employed BERT and explored two methods: span prediction and sequence labeling. Additionally, the paper evaluated several models beyond BERT, including a Convolutional Neural Network with Gated Recurrent Units (CNN-GRU) and a SVM. The SVM, trained with multi-platform data, achieved the highest F1-score on the HatEval test set, indicating its strength in cross-platform detection. However, BERT outperformed the other models in detecting hate speech on the Extremist forums and HackForums test sets, demonstrating its superior ability to capture the linguistic nuances of hate speech within these contexts.

Hate speech is a global issue that remains unexplored in many languages. In \citep{del2023socialhaterbert} authors introduces SocialHaterBERT, a  model designed for hate speech detection that is trained in both English and Spanish. This research builds upon the HaterNet algorithm by leveraging the capabilities of BERT. One of the most significant contributions of this study to the state of the art is the incorporation of social network features into the classification model. By analyzing statistics and characteristics derived from the interaction graph of Twitter users, the model enhances its ability to detect hate speech by considering not only the content of the messages but also the social context in which they are shared. The results demonstrate that SocialHaterBERT outperforms HaterBERT by 4\% and shows a 19\% improvement over the original HaterNet algorithm. These findings underscore the importance of integrating user characteristics into hate speech detection systems, as this approach allows for a more nuanced and contextual understanding of the messages being analyzed. The study concludes that the fusion of user profile data with textual analysis not only enhances the detection capabilities of hate speech models but also paves the way for new research opportunities in the field of cyber abuse. 

Authors in \citep{ramos2024leveraging} conducted an exhaustive review that systematically evaluates various transfer learning models for hate speech detection in European Portuguese, focusing particularly on BERT-based models and generative models. The study utilized two distinct annotated corpora consisting of YouTube comments and Twitter tweets to assess model performance. The findings reveal that the HateBERTimbau model achieved the highest F-score of 87.1\% on the YouTube corpus, while GPT-3.5 performed best on the Twitter corpus with an F-score of 50.2\%. The evaluation was centered on the positive classes, ensuring robustness in the results despite the absence of oversampling or undersampling techniques, demonstrating the models' efficacy in handling class imbalances within the datasets.


In \citep{garcia2023evaluating}, the authors compiled existing Spanish-language datasets, such as the MisoCorpus 2020, AMI 2018, HaterNet, and HatEval 2019, creating a valuable resource for analyzing hate speech across different contexts. The study focuses on knowledge integration strategies, which combine diverse feature sets to build more robust detection systems. Specifically, the authors proposed a method that fuses linguistic features with transformer-based models. These linguistic features include morphosyntactic, semantic, stylometric, and negation-related elements. For the detection process, they utilized Spanish BERT variants like BETO and BERTIN to improve accuracy. Additionally, the authors introduced a model resolver that integrates various feature sets, employing ensemble learning techniques to further enhance the system’s ability to detect hate speech in Spanish-language texts. This approach demonstrates the value of combining linguistic insights with advanced machine learning techniques for more nuanced hate speech detection. In \citep{garcia2023leveraging}, the authors investigate the effectiveness of Zero-Shot Learning (ZSL) and Few-Shot Learning (FSL) strategies for hate speech detection in low-resource languages, specifically focusing on Spanish applications. The study evaluates the performance of various generative models, including T5, BLOOMZ, and Llama-2, in detecting hate speech. The research explores the cross-lingual capabilities of these models, assessing their effectiveness in both English and Spanish contexts. Under ZSL conditions, the models classify hate speech without prior examples, relying on descriptive prompts of hate speech categories to infer classification based on their contextual understanding. In FSL, the models are provided with a small number of labeled examples (typically five) per hate speech category, which are included in the prompts to help the models grasp the nuances of the task. Llama-2 consistently outperformed other models, especially in English datasets, demonstrating that generative models can effectively bridge the gap between data scarcity and accurate hate speech detection. Similarly Pan et al. in \citep{pan2024comparing} examine the effectiveness of various models in detecting hate speech and cyber abuse, with a particular emphasis on sexism. The study evaluates several models, including BERT, RoBERTa, DeBERTa, mBERT, ALBERT, DistilBERT, and XLM-RoBERTa, using both fine-tuning and contextual learning approaches, such as zero-shot and few-shot learning. An in-depth error analysis is also conducted, revealing challenges like difficulties in differentiating between hate speech categories and handling figurative language.

Another significant challenge in low-resource languages is the lack of available datasets, obstructing the development of effective models for tasks like hate speech and cyberbullying detection. Recently, there has been a growing trend of research papers proposing their own datasets for these underrepresented languages, which is highly valuable for advancing research beyond English. In \citep{MontesinosCnovas2023SpanishHD}, the authors introduced a novel multi-label corpus of 7,483 manually annotated tweets, categorized into four classes: safe, aggressive, misogynistic, and racist, specifically within the context of Spanish football matches. They employed state-of-the-art pre-trained LLMs like BETO, MarIA, and multilingual BERT to develop an ensemble classifier. The model achieved an impressive macro-weighted F1-score of 88.71\%, demonstrating its effectiveness in identifying and classifying various forms of abusive content, including aggression, misogyny, and racism. The study highlighted that while individual LLMs performed well, their combination with linguistic features—such as proper nouns, contextual keywords, and psycho-linguistic elements—led to a significant improvement in classification accuracy. The authors noted that the models showed strong results across different hate speech categories, though they also identified opportunities for further enhancement by expanding the dataset and refining the classifiers to better capture a broader spectrum of cyber abuse expressions in sports-related social media content. Sreelakshmi et al. in \citep{sreelakshmi2024detection} developed an annotated CodeMix Malayalam-English hate speech corpus and conducted an in-depth study on detecting hate speech and offensive language in CodeMix texts from Dravidian languages, specifically Kannada, Malayalam, and Tamil. Their key contributions include the introduction of a cost-sensitive learning approach to address class imbalance in the datasets and a thorough evaluation of various multilingual transformer models such as BERT, DistilBERT, LaBSE, and MuRIL, alongside traditional machine learning techniques like SVM. The results demonstrated that models such as MuRIL and DistilBERT outperformed others, achieving notable accuracy and F1-scores, highlighting their effectiveness in hate speech detection in CodeMix text. Similarly to previously reviewed works, Almaliki et al. in \citep{almaliki2023abmm} introduced a new dataset comprising approximately 10,000 annotated tweets specifically for hate speech detection in the Arabic language. To evaluate the dataset's effectiveness, they conducted experiments using BERT models. Urdu has garnered significant attention in recent years, with researchers increasingly focusing on the language's unique nuances.  Ali et al. \citep{ali2022hate} contributed to this growing body of work by compiling a dataset of around 20,000 tweets using the open-source tool Twint, specifically targeting hate speech-related keywords. The study explores the effectiveness of transfer learning techniques, utilizing pre-trained transformer models like BERT to tackle the challenges of hate speech detection in Urdu. In a similar approach, Arshad et al. \citep{arshad2023uhated} present their work on hate speech detection. This study leverages transfer learning techniques woth RoBERTa, specifically tailored for Urdu, to address the challenges associated with detecting hate speech in this underrepresented language. Also in Urdu, \citep{bilal2023roman} introduced a new dataset comprising 173,714 Roman Urdu hate speech examples and conducted a comparative analysis between deep learning and transformer models on this dataset.

Nandi et al. also highlight the potential of using pre-trained language models to construct an ensemble method for hate speech classification, particularly focusing on low-resource Indian languages \citep{nandi2024combining}. The study introduces a feature fusion-based ensemble model designed to detect hate speech in Bengali, Marathi, and Hindi. The approach leverages two pre-trained language models: multilingual BERT (mBERT) and IndicBERT. The features extracted from both models are concatenated, and a traditional softmax layer is employed to perform the final classification. This fusion of feature representation proved effective, with the model achieving weighted F1 scores of 0.923 for Bengali, 0.815 for Hindi, and 0.924 for Marathi, significantly outperforming existing benchmarks for the Bengali and Marathi datasets. For hate speech detection in Bengali, \citep{keya2023g} proposed a BERT architecture combined with Gated Recurrent Units, a type of recurrent neural network that is particularly well-suited for handling sequential data such as text. 

In \citep{arcila2022detect}, Arcila et al. introduced an ad hoc dataset manually annotated for the detection of hate speech and racist messages, specifically focusing on topics related to migration and refugees. The dataset primarily comprises Spanish-language messages but has been extended to include messages in Greek and Italian as well, making it more comprehensive in its coverage of multilingual hate speech. The authors conducted a comparative analysis of traditional data mining techniques, deep learning architectures, and transformer-based models for binary classification of hate speech. The results indicated that deep learning and transformer models significantly outperformed traditional algorithms, demonstrating superior performance metrics in detecting hate speech across different languages. The study emphasizes the importance of developing high-quality, validated datasets as a fundamental step for training effective and reliable predictive models in the domain of hate speech detection. This work underscores the critical role that diverse, annotated datasets play in enhancing the accuracy and generalizability of machine learning models across multilingual contexts.

In \citep{maity2024hatethaisent}, the authors proposed a novel tagged dataset for hate speech detection in the Thai language. A key contribution of this work is the incorporation of sentiment analysis through a systematic annotation process, where each social media post is labeled with both hate speech and sentiment information. To classify hate speech, the study employs a dual-channel multitask model called ThaiCaps, which integrates both BERT and FastText embeddings. BERT effectively captures contextual information and semantic nuances, while FastText offers character-level features, enhancing the model's robustness against the noisy nature of social media data. This multitask approach enables the model to simultaneously tackle hate speech detection and sentiment analysis, leveraging the interrelated nature of these tasks to improve overall performance. The results from the ThaiCaps model are impressive, achieving an overall accuracy of 89.67\% and a macro F1 score of 89.79\% for hate speech detection, alongside accuracy and F1 scores of 80.92\% and 80.97\% for sentiment detection, respectively. This highlights the model's effectiveness in addressing both hate speech and sentiment analysis within the Thai language context.

In the context of Arabic language hate speech detection, Althobaiti et al. \citep{althobaiti2022bert} and Husain et al. \citep{husain2022investigating} explored the use of BERT architectures to enhance performance in this area. Husain et al. \citep{husain2022investigating} conducted a comparative analysis between traditional models and BERT, emphasizing the importance and different approaches to preprocessing for improving hate speech detection. Their study revealed significant variations in classifier performance based on the preprocessing techniques employed. While traditional classifiers like SVM demonstrated better results with fully processed text, BERT-based models excelled with targeted preprocessing methods, such as emoji conversion and dialect normalization. Althobaiti et al. \citep{althobaiti2022bert} proposed a BERT-based classification system that integrates sentiment analysis features and emoji transcriptions into the model's feature set. The incorporation of sentiment features led to a modest improvement in the model's overall performance. Additionally, the impact of emoji transcriptions on the model's effectiveness varied depending on the size of the annotated dataset and the uniqueness of the emojis as distinguishing features. The potential of incorporating emotional features and tone analysis is further explored in \citep{anjum2024hatedetector}, where the authors propose a multilingual hate speech detection model that combines BERT with a Multilayer Perceptron (MLP) and a Logistic Regression classifier. This approach aims to effectively analyze hate speech across multiple languages. A key feature of their system is the incorporation of a profanity detection technique, which leverages a curated list of profane words to assess the emotional tone of tweets, thereby enhancing the model's ability to accurately identify hate speech.

In \citep{pookpanich2024offensive}, the authors investigated the impact of hate speech in live-streaming messages during football matches, focusing specifically on detecting offensive language in Thai YouTube live chat. The study utilized five transformer-based language models: BERT, XLM-RoBERTa, DistilBERT, WangchanBERTa, and TwHIN-BERT. They carried out an implementation of a two-stage labeling process, combining automated and manual labeling to enhance the quality of the dataset, which consists of over 2 million messages categorized as neutral, positive, negative, or hate speech. The authors placed special emphasis on identifying negative messages with offensive content, paying attention to the intricacies of Thai slang and colloquial language, which are often indicators of hate speech. To address the challenge of class imbalance, they introduced "fake hate speech" messages—containing hate speech keywords without necessarily being abusive—helping to balance the dataset and improve model training. Results showed that WangchanBERTa, in particular, achieved the best performance, with high accuracy, precision, recall, and F1-scores, highlighting its effectiveness in detecting nuanced hate speech in the Thai language.


Most studies on hate speech detection rely on deep learning-based classification models, which often function as ``black boxes'' with limited interpretability. In \citep{mehta2022social}, the authors take a different approach by focusing on model explainability. They apply interpretability techniques such as SHAP (Shapley Additive Explanations) \citep{lundberg2017unified} values and LIME (Local Interpretable Model-agnostic Explanations) \citep{ribeiro2016should} to shed light on the decision-making processes of various models. LIME, in particular, is applied to predictions made by both traditional classifiers and deep learning models like BERT variants. It generates local explanations by assigning weights to individual words, revealing how much each word contributes to the model's decision on whether a comment contains hate speech. The study evaluates a combination of traditional machine learning techniques alongside more advanced deep learning architectures, such as BERT integrated with different kinds of traditional neural netwokrs architecture. The results show that BERT-based models significantly outperform traditional classifiers, achieving accuracies of up to 93.67\%. By employing explainability techniques like LIME, the authors not only enhance the performance of hate speech detection models but also make them more transparent, offering valuable insights into how specific words or phrases influence the predictions. This approach bridges the gap between model accuracy and interpretability, making the models more trustworthy and actionable.


In \citep{saleh2023detection}, Saleh et al. approached hate speech detection from two distinct approaches: one utilizing BERT and the other focusing on constructing an extensive hate speech-specific word embedding to serve as features for a biLSTM classifier. The primary contribution of the paper is the creation of this comprehensive word embedding, which was trained on a curated dataset comprising approximately 1 million sentences sourced from hate speech domains on Twitter. By building the embedding in an unsupervised manner, the authors provide a valuable resource that captures the nuances and context of hate speech language. The results demonstrated that this method, which combines unsupervised word embedding with a biLSTM classifier, is highly effective, achieving state-of-the-art performance. The study underscores the potential of leveraging unsupervised techniques in the hate speech detection domain, suggesting that they can be competitive.


The effectiveness and necessity of unsupervised techniques in understanding hate speech are explored in Ollagnier et al. \citep{ollagnier2023unsupervised}, where the authors conduct a fine-grained analysis of hate speech through multiple perspectives, including the type of hate, its directness, and target. They undertake community detection and clustering experiments using a novel approach that models cyberbullying as a graph structure. The process begins with generating embeddings using mBER and multilingual Universal Sentence Encoder (mUSE). They then compute similarities between the embeddings to construct a similarity matrix, from which a graph is built with tweets as nodes and edges representing the similarity scores. This graph serves as the basis for community analysis and clustering, employing Multi-view Spectral Clustering. The resulting clusters provide valuable insights into the different forms of hate speech, their interactions, and how they manifest across various contexts.The importance and effectiveness of hybrid systems that combine unsupervised and supervised learning methods are also emphasized by Altinel et al. \citep{altinel2024so}. In their work, the authors developed a system that integrates K-means clustering, BERT, and a textGCN classification layer for hate speech detection. In this hybrid approach, the test set is first partitioned into k subclusters using the K-means algorithm. Subsequently, for each distinct subset identified by the features resulting from the K-means division, the textGCN technique is applied. Additionally, the authors introduced a manually labeled dataset for hate speech in the Turkish language, further contributing to research in this domain.

Finally, in \citep{chhikaraidentification}, the authors delve into the issue of automated accounts spreading hate speech, focusing on identifying and differentiating AI-generated hate speech from genuine human-generated content. They introduce a Giant Language Model Test Room (GLTR), a tool designed to analyze and detect the authenticity of text produced by large language models. The tool combines BERT and GPT-2 to effectively tackle this challenge. BERT is fine-tuned on labeled datasets to classify individual words and phrases, enabling it to detect instances of hate speech based on contextual meanings. Meanwhile, GPT-2 is employed to generate synthetic examples of hate speech, which are then used to augment the training data, enhancing the model's ability to recognize various patterns. The outputs of BERT and GPT-2 are combined to evaluate how effectively generative hate speech examples can be detected. This innovative approach provides valuable insights into the challenge of identifying AI-generated hate speech, a topic that will be further explored in Section\ref{sec:generation}.

\subsubsection{Quantitative Analysis of Hate Speech detection leveraging LLMs}

In this quantitative analysis (Table \ref{tab:summary_studies}), we investigate the prevalence of various language models employed in the reviewed papers, alongside their approaches to handling unbalanced data.

The predominant language model identified across the studies is BERT and its variants, including BETO, DistilBERT, mBERT, and BERTIN. These models appear in approximately 77\% of the papers, underscoring BERT’s status as the leading choice for numerous natural language processing tasks. Furthermore, RoBERTa and its derivatives are utilized in around 14\% of the papers. LLMs such as GPT-2, FLAN-T5, and LLAMA-2 are also present, though to a lesser extent, being mentioned in approximately 28\% of the studies. This trend suggests that BERT continues to be the most widespread language model, particularly excelling in hate speech detection. Notably, it is frequently combined with LSTM models for classification tasks. The high adoption rate of BERT is further influenced by the nature of recent publications, which predominantly focus on fine-tuning for various languages, making BERT and its variants particularly suitable for fine-tuning and transfer learning applications.

Regarding the handling of unbalanced data, the approaches varied significantly among the papers. In 31.4\% of the cases, researchers successfully addressed the imbalance using techniques such as data augmentation—often employing generative AI to create synthetic examples—oversampling, or the application of weighted metrics. However, the majority of the papers (51.4\%) did not adequately address the issue of unbalanced data, despite some acknowledgment of its presence. Additionally, in 17.1\% of the papers, the issue was only partially handled, meaning that while certain methods were discussed or acknowledged, they were not comprehensively implemented or evaluated in practice.

Among the papers that did address unbalanced datasets, a variety of techniques were employed. Weighted metrics, such as weighted F1 scores, emerged as the most commonly utilized method, appearing in 4 papers. Oversampling and undersampling strategies were applied in 3 papers, while data augmentation techniques were employed in 2 studies. Furthermore, cost-sensitive learning and instance weighting were utilized in a few cases, with 2 papers implementing cost-sensitive learning and 1 paper employing instance weighting. These techniques are vital for mitigating bias toward the majority class, thereby enhancing the model's performance on predictions for minority classes.

\renewcommand{\arraystretch}{1.5} 

\begin{longtable}{p{3cm}p{3.5cm}p{3.5cm}p{4.5cm}} 
\caption{Summary of hate speech detection studies using LLMs} \label{tab:summary_studies} \\ 
\hline
\textbf{CITE (Author, Year)} & \textbf{LM or LLM Used?} & \textbf{Traditional Techniques Used?} & \textbf{Unbalanced Data Handling} \\ 
\hline
\endfirsthead

\hline
\textbf{CITE (Author, Year)} & \textbf{LM or LLM Used?} & \textbf{Traditional Techniques Used?} & \textbf{Unbalanced Data Handling} \\ 
\hline
\endhead

\hline \multicolumn{4}{r}{{Continued on next page}} \\ 
\endfoot

\hline
\endlastfoot

 \citep{ansari2024data} & BERT, bi-LSTMs & CNN & Addressed with data augmentation techniques \\ 

\citep{zhou2023automated} & BERT & SVM, CNN-GRU & Not handled \\ 

\citep{garcia2023evaluating} & BERT-based: BERTIN, BETO, BNE, M-BERT & None & Discussed but not implemented \\ 

\citep{garcia2023leveraging} & FLAN-T5, LLAMA-2, mT0, Flan-Alpaca, ALBETO, BETO, DistilBETO, mBERT, MarIA, mDeBERTa, twHIN & None & Not handled but evaluated from a binary perspective \\ 

 \citep{MontesinosCnovas2023SpanishHD} & BETO, MarIA, BERTIN, DistilBETO, AlBETO, Multilingual BERT, mBERT, XLM, Multilingual DeBERTa, mDeBERTa, TwHIN-BERT & None & Partially handled using weighted F1-Score \\ 

 \citep{sreelakshmi2024detection} & BERT, DistilBERT, LaBSE, MuRIL, XLM, IndicBERT, FNET & SVM, Random Forest, Naive Bayes, KNN, Logistic Regression & Handled using cost-sensitive learning \\ 

 \citep{su2023ssl} & RoBERTA & GAN & Not handled \\ 

 \citep{del2023socialhaterbert} & BERT, BETO & None & Not handled but well evaluated \\ 

\citep{pookpanich2024offensive} & BERT, XLM-RoBERTa, DistilBERT, WangchanBERTa, TwHIN-BERT & None & Correctly handled using oversampling \\ 

\citep{mehta2022social} & BERT + ANN, BERT + MLP, LSTM & Decision Trees, Random Forest, Logistic Regression, Naïve Bayes & Handled using weight optimization \\ 

\citep{arcila2022detect} & BERT & NB, Logistic regression, SVM, RNN & Handled via instance weighting and macro-F1/AUC \\ 

\citep{maity2024hatethaisent} & BERT, bi-LSTMs & FastText & Not handled \\ 

\citep{anjum2024hatedetector} & BERT & MLP, Logistic Regression, BoW & Not handled \\ 

\citep{nandi2024combining} & BERT (mBERT) and Indic BERT & None & Not handled, evaluated with multiple F1 measures \\ 

\citep{putra2024semi} & BERT & CNN & Correctly handled using class importance \\ 

\citep{yun2023bert} & Soongsil-BERT, KcELECTRA, BERT, RoBERTa & XGBoost & Correctly handled with weighting and oversampling \\ 

\citep{chhikaraidentification} & BERT, GPT-2 & None & Handled with data augmentation \\ 

\citep{saleh2023detection} & BERT, bi-LSTMs & Word2Vec & Well evaluated with weighted metrics \\ 

\citep{almaliki2023abmm} & BERT & None & Not handled but evaluated by class F1 \\ 

\citep{ramos2024leveraging} & BERTimbau, BERTimbau-hatebr, mDeBERTa-hatebr, HateBERTimbau, GPT-3.5, Gemini-Pro, Mistral-7B-Instruct-v0.3, CNN, LSTM & None & Correctly evaluated with emphasis on positive F-score \\ 

\citep{malik2024hate} & Urdu-RoBERTa, Urdu-DistilBERT & None & Not handled, only weighted F1 used \\ 

\citep{pan2024comparing} & Mistral-7B-Instruct, Zephyr-7B, StableBeluga-7B, Tulu-2, LlaMa-2, Mistral-7B-v0.1, BERT, RoBERTa, DeBERTa, mBERT, ALBERT, DistilBERT, XLM-RoBERTa & None & Evaluated with macro and weighted F1, not handled \\ 

\citep{ollagnier2023unsupervised} & mBERT, mUSE & Multi-view Spectral Clustering & Not handled, dataset adjusted by removing labels \\ 

\citep{arshad2023uhated} & RoBERTa & FastText & Not handled \\ 

\citep{bilal2023roman} & BiLSTM, BERT & Logistic Regression, XGBoost, Random Forest, CNN & Not handled \\ 

\citep{ali2022hate} & BERT, DistilBERT & FastText + BiGRU & Not handled, evaluated with macro-F1 \\ 

\citep{firmino2024improving} & BERTimbau, Italian BERT, BERT, XLM-R & None & Correctly handled with extensive evaluation \\ 

\citep{toliyat2022asian} & BERT, bi-LSTMs & NB, KKN, XGBoost, Random Forest & Correctly handled and evaluated \\ 

\citep{khan2022bichat} & BERT, bi-LSTMs & CNN & Correctly evaluated on balanced/unbalanced configurations \\ 

\citep{althobaiti2022bert} & BERT & SVM, Logistic Regression & Partially handled with synthetic datasets \\ 

\citep{karayiugit2022homophobic} & M-BERT & None & Correctly handled using oversampling/undersampling \\ 

\citep{keya2023g} & BERT, GRU & None & Not handled  \\ 

\citep{gupta2021bert} & BERT & None & Partially Handled (acknowledged issue of imbalance) \\ 

\citep{wullach2022character} & BiGRU, GPT-2 & None & Handled with data augmentation \\ 

\citep{altinel2024so} & BERT & TextGCN, Kmeans, Word2Vec, Doc2Vec, Glove, TF &  Partially Handled (acknowledged issue of imbalance) \\ 

\citep{mazari2024bert} & BERT, bi-LSTMS, BiGRU & MLP & Correctly evaluated with macro-averaged measures \\ 

\citep{husain2022investigating}  & AraBERT, ArabicBert, LSTM   & SVM, Bagging, RF, Logistic Regression, RNN & Correctly evaluated with macro-averaged measures \\ 
\hline
\end{longtable}

\subsection{Cyberbullying detection}
\label{sec:cyber-detection}

With the exponential growth of internet users and the increasing prevalence of social media, harmful behaviors like cyberbullying have become widespread. Detecting cyberbullying, however, remains a challenging task due to the linguistic subtleties and contextual nuances often embedded in offensive language. Traditional approaches rely heavily on extensive human-annotated datasets, which are both time-consuming and costly to produce. In an effort to overcome these limitations, \citep{al2024towards} proposes a novel approach using Deep Contrastive Self-Supervised Learning (DCSSL). This method automates data augmentation and labeling, enabling the training of models on larger datasets without the need for extensive manual intervention. By starting with a small, labeled dataset, the model identifies similar examples in unlabeled data, effectively generating new instances that capture the complex nature of cyberbullying language. To automate the labeling process, a fine-tuned sentence encoder is used to compute sentence similarities, assigning pseudo-labels to unlabeled data based on their resemblance to labeled examples. The model combines BERT with a Bi-LSTM network to leverage the strengths of both architectures, resulting in richer sentence representations. The DCSSL model achieves a macro-averaged F1 score of 0.9231, significantly outperforming baseline models in detecting cyberbullying.

Has we shown in section \ref{sec:forms_of_textual_cyberabuse} cyberbullying is closely linked to significant psychological and emotional distress. In this context, Maity et al. \citep{maity2022multitask} approach the issue of cyberbullying from the perspective of sentiment and emotional analysis. They point out a notable gap in existing research, which often overlooks the integration of sentiment and emotion data for detecting cyberbullying—especially in code-mixed languages like Hindi and English. To address this, the authors developed a multitask, multimodal framework that incorporates both sentiment and emotion analysis to enhance the detection of cyberbullying. Their contribution includes the creation of a novel dataset, BullySentEmo, annotated with labels for cyberbullying, sentiment, and emotion. The proposed framework utilizes BERT and VecMap embeddings along with a bi-GRU model to analyze both text and emoji modalities. The results are impressive, with accuracy rates of 82.05\% for cyberbullying detection, 77.87\% for sentiment analysis, and 58.05\% for emotion recognition. The study underscores the crucial role of sentiment and emotional cues in cyberbullying detection, revealing that posts labeled as bullying tend to exhibit negative sentiments and emotions such as anger and disgust. Also focusing on emotional features within a privacy-preserving framework, \citep{samee2023safeguarding} introduces an innovative federated architecture aimed at enhancing both the security and accuracy of cyberbullying detection models. A key contribution of the study is the introduction of eight novel emotional features extracted from textual tweets, which provide deeper insights into the emotional undertones of messages, thereby improving the detection of cyberbullying instances. The authors address the challenge of imbalanced data by employing data undersampling techniques before distributing the data to each federated node for analysis. The results demonstrate that the proposed framework, particularly through the integration of the BERT model, not only improves detection accuracy but also ensures user privacy via federated learning, outperforming traditional centralized methods in cyberbullying identification.

Cyberbullying can manifest in various forms, particularly in online platforms like Twitter, where the role of bystanders can significantly influence the nature of the bullying. In \citep{alfurayj2024chained}, this complexity is explored by introducing a dataset that incorporates bystander roles as a feature in cyberbullying detection, enabling a more nuanced analysis of interactions. Bystanders are categorized into defenders, instigators, neutrals, and others. The proposed model utilizes a novel chained architecture consisting of two sequential classification layers. The first layer classifies bystander roles, while the second layer detects cyberbullying, allowing the model to capture the interdependencies between bystander behavior and bullying. The authors fine-tune a pre-trained BERT model, enhanced with LSTM networks, to extract textual features. The model achieves a significant F1-score of 89\% in classifying bystander roles, marking a 25.35\% performance improvement over traditional methods. While the paper acknowledges the challenge of unbalanced datasets, no specific solutions for this issue are provided. However, they propose the use of weighted metrics to analyze the results, yielding more robust evaluations in unbalanced scenarios.


Recent trends in social media interactions increasingly involve the use of text overlaid on images, creating new opportunities for cyberbullying to emerge in the multimodal domain of social networks. In their work, Li et al. \citep{li2024integrating} address this issue by developing a multimodal system for cyberbullying detection, leveraging a dataset consisting of 5,211 irony-aware cyberbullying text-image pairs and 6,598 non-cyberbullying pairs. This dataset enables a detailed analysis of the subtleties in cyberbullying expression. The proposed model integrates cutting-edge techniques such as Graph Isomorphism Networks (GIN) for feature transformation, BERT for text processing, and Vision Transformer (ViT) for image analysis. By combining these methods, the multimodal approach enhances the structural information embedded in the data, allowing for a more nuanced understanding of cyber abuse behaviors. The results demonstrate that the framework significantly outperforms traditional single-modality approaches, whether based solely on text classification or image recognition, particularly in detecting irony-aware cyberbullying. Additionally, the model reduces decision-making bias through a soft voting mechanism that aggregates predictions from various feature combinations, improving accuracy and robustness. The study also effectively addresses the challenges posed by the imbalanced dataset using the Synthetic Minority Oversampling Technique (SMOTE) \citep{chawla2002smote,suhas2022novel}. Performance was evaluated through metrics such as weighted F1 and AUC scores.


In the multilingual domain, \citep{razi2024multilingual} explores linguistic variations in text messages written in Urdu, Roman Urdu, and English. The research focuses on developing a comprehensive dataset that captures these variations, with a particular emphasis on cyberbullying instances characterized by aggression, repetition, and intent to harm, key elements for a deeper understanding of cyber abuse. The authors propose a novel framework for classifying text messages as aggressive or non-aggressive, leveraging fine-tuned models like m-BERT and MuRIL to improve detection accuracy. The approach demonstrates strong performance, achieving a precision of 0.93, recall of 0.92, and an F1-score of 0.92, showcasing its effectiveness in identifying cyberbullying. However, while the results are promising, the dataset is somewhat imbalanced, and no techniques like undersampling or oversampling were applied to address this issue. The model’s strength lies in its ability to capture the core aspects of cyber abuse namely aggression, repetition, and intent to harm, allowing for a more comprehensive classification of cyberbullying incidents.

\citep{mahmud2024exhaustive} presents a dataset of over 5,000 manually curated text samples from social media platforms, focusing on the Bangla and Chittagonian languages. The study offers a comprehensive comparison of popular deep learning models, transformer-based architectures, and traditional machine learning techniques for text classification. The results highlight that deep learning models, especially CNNs and transformer-based architectures, significantly outperform traditional machine learning approaches. These advanced models achieved accuracies ranging from 0.69 to 0.811, showcasing their superior ability to handle the complexities of multilingual and context-specific cyberbullying detection in these languages. Similarly, \citep{saini2023enhancing} presents findings where the authors compare an ensemble method, combining a Convolutional Neural Network (CNN) with a Support Vector Machine (SVM), against a BERT-based model for cyberbullying detection. Their results highlight that while the CNN–SVM ensemble achieved a strong accuracy of 96.88\%, the BERT model outperformed it, achieving a higher accuracy of 97.34\%, showcasing the superior capability of transformer-based models in handling complex language tasks. The effectiveness of transformers is further highlighted in \citep{mahdi2024enhancing}, where the authors demonstrate the significant impact of fine-tuning a BERT model specifically for Arabic languages in the context of cyberbullying classification. The study shows how this targeted fine-tuning leads to remarkable improvements over baseline models, showcasing the model's ability to capture the unique linguistic characteristics of Arabic and significantly enhance classification performance.

In multilingual societies like India, code-switching—mixing multiple languages within the same conversation—is prevalent, and \citep{paul2023covid} addresses this by curating a new annotated dataset of over 22,000 tweets that blend English and Hindi, specifically focusing on instances of cyberbullying. Sayanta et al. proposed a powerful ensemble model that integrates both machine learning and deep learning techniques to improve detection accuracy. This model combines traditional classifiers, such as SVM and Logistic Regression, with advanced deep learning architectures like Multilayer Perceptron (MLP), CNN, BiLSTM, and BERT. For embeddings, the study utilized methods like FastText, CBOW, and TF-IDF vectorization to build domain-specific representations. The proposed deep ensemble model achieved A macro-averaged F1 score of 0.93, demonstrating state-of-the-art performance in identifying cyberbullying in the code-switched dataset. The authors emphasized the critical role of language-specific word embeddings and their concatenation in effectively processing code-switched text, enabling the model to better capture the linguistic nuances present in cyberbullying. However, the study did not account for unbalanced data scenarios, and while the performance is strong, certain cyberbullying techniques are underrepresented, potentially impacting the model’s generalization to more complex forms of abuse.


Continuing with ensemble techniques \citep{muneer2023cyberbullying} introduces an approach to detecting cyberbullying on social media platforms, specifically targeting Twitter (now X) and Facebook. The study's key contribution is the development of a stacking ensemble learning model that integrates several deep learning techniques, including a fine-tuned BERT model adapted for cyberbullying data, along with biLSTMs and CNNs. This combination enhances the accuracy and robustness of cyberbullying detection. The research utilizes labeled text datasets from both Twitter and Facebook, focusing on offensive and non-offensive comments to train the models in distinguishing cyberbullying from benign interactions. The stacking ensemble model demonstrates strong performance, achieving an accuracy of 97.4\% on the Twitter dataset and 90.97\% on a combined dataset of Twitter and Facebook. Furthermore, the study highlights notable improvements in processing efficiency, with the proposed model requiring only 2 minutes and 45 seconds for data processing, significantly faster than baseline models. The authors address the challenge of imbalanced data, which is prevalent in cyberbullying detection tasks, by employing robust evaluation metrics such as F1-score and AUC, well-suited for such scenarios. Notably, despite being older techniques, the paper effectively leverages CBOW and Word2Vec embeddings, achieving impressive results in the context of cyberbullying detection.

In \citep{ahmed2022performance} Ahmed et al presented a comprehensive analysis of transformer-based architectures for detecting trait-based cyberbullying in social media comments, contributing to the development of automated detection systems. The authors evaluate several well-known models, including BERT, DistilBERT, RoBERTa, XLNet, and GPT-2, highlighting their effectiveness in capturing the nuances of text data. The study utilizes two publicly available datasets: the Fine-Grained Cyberbullying Dataset (FGCD), which is balanced and categorizes comments into six distinct classes, and the Twitter-parsed Cyberbullying Dataset, which is imbalanced and includes three classes. To address the challenges posed by imbalanced data, the authors implement a semi-supervised learning technique called Dynamic Query Expansion to enhance the representation of underrepresented classes.

In an effort to maximize the benefits of combining multiple models, \citep{chen2024chinese} integrates two distinct architectures: XLNet and Deep BiLSTM. This approach exemplifies model fusion, where the strengths of each architecture are leveraged to enhance the performance of cyberbullying detection. The study utilizes the COLDATASET, which was meticulously constructed by augmenting an existing Chinese offensive language dataset with 1.66k offensive remarks from real cyberbullying incidents and one-star reviews sourced from the Douban platform. This enriched dataset captures both explicit and implicit forms of cyberbullying, which are essential for effective detection. The results show that the proposed hybrid model outperforms traditional machine learning and other deep learning approaches in detecting cyberbullying. The use of pre-trained LLMs like XLNet for representation learning proves especially beneficial. In the classification phase, the model addresses data imbalance through oversampling techniques, allowing it to learn implicit and explicit features of cyberbullying more effectively. While the dataset expansion improves balance, it is worth noting that metrics such as AUC, crucial for evaluating performance in imbalanced scenarios, are missing.


\citep{ferrer2024using} contributes to the field of cyberbullying detection by developing an AI-based virtual companion aimed at helping adolescents, particularly those with autism, recognize and respond to instances of cyberbullying. The research leverages a combination of datasets, including the limited KIDS-cyberbullying dataset and a larger Kaggle cyberbullying dataset, which consolidates diverse examples of online harassment from various social media platforms. To address the imbalanced nature of the datasets, data augmentation techniques such as paraphrasing and back-translation were employed. The results demonstrate that lightweight models, specifically T5-small and MobileBERT, are capable of effectively detecting cyberbullying in real time within a web-based system. The study emphasizes the importance of considering context in cyberbullying detection, as traditional approaches that focus solely on profanity often miss more subtle forms of abuse. The paper concludes by highlighting the potential of AI-based virtual companions in educating adolescents about cyberbullying and improving their ability to identify and cope with such harmful behaviors.

\subsubsection{Quantitative Analysis of Cyberbullying detection leveraging LLMs}

In Table \ref{tab:cyberbullying_detection}, we present a quantitative analysis of cyberbullying detection, paralleling the analysis conducted for hate speech.

The most commonly used language models in the new set of studies are BERT and its variants, which appear in approximately 61.5\% of the papers. This trend is consistent with the findings from Table \ref{tab:summary_studies}, where BERT also dominated, appearing in 77\% of the studies. Other models such as RoBERTa and XLNet are represented in a smaller proportion of the new studies, with RoBERTa appearing in 7.7\% of the papers. This suggests that BERT continues to be a prevalent choice for researchers in cyberbullying. 

In the context of handling unbalanced data, there is a notable diversity in the approaches employed across the papers. 38.5\% of the studies effectively handled unbalanced datasets using methods such as data augmentation and oversampling techniques. This is a positive increase compared to Table \ref{tab:summary_studies}, where only 31.4\% of the papers successfully addressed the imbalance. Specifically, data augmentation techniques were utilized in 2 papers, while effective balancing was achieved through over-sampling and under-sampling in another study. Furthermore, 30.8\% of the papers acknowledged the issue of unbalanced data but only partially addressed it. This aligns with the previous analysis, where 17.1\% of the studies reported partial handling of unbalanced datasets.

Among the studies that addressed unbalanced data effectively, various techniques were employed. The most commonly used approach in the new table involved data augmentation, applied in 2 studies, while 2 papers reported using over-sampling and under-sampling methods. Weighted metrics, particularly in the context of partial handling, were acknowledged in 2 studies. The effective handling of unbalanced data using methods such as SMOTE was implemented in one paper, highlighting the growing recognition of the importance of addressing this issue in model training.

The findings from the reviewed papers on cyberbullying reveal trends similar to those observed in hate speech detection. BERT remains the dominant language model; however, while the handling of unbalanced data shows improvement, significant challenges persist, with a considerable proportion of studies still failing to address the issue comprehensively.

\begin{table}[]
\centering
\small
\renewcommand{\arraystretch}{1.5} 
\begin{adjustbox}{max width=\textwidth}
\begin{tabular}{p{4.5cm}p{3.5cm}p{3.5cm}p{3.5cm}} 
\hline
\textbf{CITE (Author, Year)} & \textbf{LM or LLM Used?} & \textbf{Traditional Techniques Used?} & \textbf{Unbalanced Data Correctly Processed?} \\ 
\hline
Razi et al., 2024 \citep{razi2024multilingual} & m-BERT, muRIL & None & Not Handled \\ 

Ferrer et al., 2024 \citep{ferrer2024using} & T5-small, MobileBERT & None & Handled with data augmentation techniques) \\ 

Muneer et al., 2023 \citep{muneer2023cyberbullying} & BERT, bi-LSTMs & CBOW, Word2Vec & Partially Handled (acknowledged issue of imbalance) \\ 
Chen
et al., 2024 \citep{chen2024chinese} & XLNet, bi-LSTM & None & Handled Effectively (balanced dataset via over/under-sampling) \\ 

Paul et al., 2023 \citep{paul2023covid} & BERT, bi-LSTMs & SVM, logistic regression, CBOW, FastText, TF-IDF & Not Handled \\ 

Mahmud et al., 2024 \citep{mahmud2024exhaustive} & BERT, XLM-Roberta, BiLSTM, CNN, GRU & SVM, logistic regression, RF, KNN, NB & Not Handled \\ 

Li et al., 2024 \citep{li2024integrating} & BERT, Graph Isomorphism Networks & XGB, RF, LR, KNN, ML, LGBM & Handled  (SMOTE applied) \\ 

Alfurayj et al., 2024 \citep{alfurayj2024chained} & BERT, bi-LSTMs & None & Partially Handled (evaluated with weighted metrics) \\ 

Al et al., 2024 \citep{al2024towards} & BERT, bi-LSTMs, SimCSE & None & Handled with data augmentation techniques \\ 

Maity et al., 2022 \citep{maity2022multitask} & BERT, bi-GRU & VecMap & Not Handled \\ 

Mahdi et al., 2024 \citep{mahdi2024enhancing} & BERT & None & Not Handled \\ 

Samee et al., 2023 \citep{samee2023safeguarding} & BERT & None & Handled Effectively (under-sampling applied) \\ 

Saini et al., 2023 \citep{saini2023enhancing} & BERT & SVM + CNN Ensemble & Partially Handled (addressed via ROC curves) \\ 

Alrowais et al., 2024 \citep{alrowais2024robertanet} & RoBERTa & GloVe & Not Handled \\ 

Ahmed et al., 2022 \citep{ahmed2022performance} & BERT, GPT-2, XLNet, RoBERTa, DistilBERT & SVM, XGBoost & Handled Effectively (Dynamic Query Expansion) \\ 
\hline
\end{tabular}
\end{adjustbox}
\caption{Overview of Cyberbullying Detection Research}
\label{tab:cyberbullying_detection}
\end{table}

\subsection{Generation of Abusive Content}
\label{sec:generation}

Before concluding, it is important to acknowledge that LLMs are also being misused for malicious purposes. While this aspect of artificial intelligence is less visible, a review of the literature reveals several papers discussing how LLMs are being exploited to facilitate cyber abuse and other harmful activities on social media. It is crucial to note that this section offers a perspective on the broader implications of LLMs and falls outside the scope of our primary survey. 

In \citep{divyatransforming}, the authors acknowledge the challenges associated with the misuse of generative AI, particularly in harmful contexts such as the creation of fake news and disinformation. The realistic content generated by AI poses significant risks, as it can be exploited for social engineering and manipulation. Moreover, generative AI’s capacity to produce highly realistic images and videos, such as deepfakes, raises additional concerns around identity theft and privacy invasion. Similarly, \citep{wach2023dark} discusses several controversies, threats, and drawbacks associated with generative AI, including issues such as algorithmic bias and violations of personal data privacy. The authors explicitly note that ``\textit{Generative AI can be used for cyberbullying or other forms of online harassment}''. They explain that by employing AI to create highly personalized attacks, individuals or groups can leverage social surveillance to identify vulnerable targets and initiate precisely targeted campaigns of harassment or abuse.

In the realm of deepfakes—images and videos generated by generative AI that often closely resemble reality—\citep{romero2024generative} examines their implications in relation to the EU's Artificial Intelligence Act. The article underscores the necessity of a balanced approach that protects fundamental rights while addressing the risks associated with deepfakes, including misinformation \citep{barari2021political}, extortion \citep{blancaflor2024deepfake}, and privacy violations \citep{dobrobaba2022deepfakes}.

In the realm of text generation, \citep{bullock2019automated} conducts an insightful review of the risks associated with using generative AI for political content, particularly its potential for societal disinformation. The authors identify several risks, including the automated generation of disinformation, fake news dissemination, hate speech, and impersonation. To explore these risks, they conduct a proof-of-concept experiment that illustrates how easily AI can generate politically sensitive text, emphasizing the implications for political stability, peace, and security. The study uses English-language transcripts of speeches delivered by political leaders at the United Nations General Assembly (UNGA) from 1970 to 2015 as training data. Leveraging the AWD-LSTM model, they generated text by 'seeding' it with the start of a sentence or paragraph. The quality of the generated text was evaluated based on its coherence and similarity to human-produced content. The results were notable: high-quality text for benign topics was generated around 90\% of the time, while generating more inflammatory or sensitive speech required multiple attempts, with approximately 60\% success in producing acceptable outputs. The findings highlight both the potential and the challenges of using AI to generate politically charged content.

One of the most interesting studies in this field is \citep{marcondes2024exploratory}. This paper reveals that troll factories likely utilize a hybrid approach, integrating both human oversight and automated tools, rather than relying solely on automation. The study investigates the application of LLMs for generating persuasive content on social media, demonstrating their effectiveness in producing contextually relevant tweets. Additionally, the research introduced a General Twitter Corpus to improve LLM performance. The findings suggest that while LLMs can significantly contribute to the dissemination of misinformation and trolling behaviors, human involvement remains crucial for the effective management and coordination of troll operations.

Conversely, we identified a paper that highlights the potential of generative models for creating counter-speech, which serves to respond to hate speech and mitigate the spread of harmful content. Specifically, \citep{pranesh2021towards} proposes a data-driven natural language generation task aimed at automatically generating counter-narratives in response to online hate speech. This research utilizes three state-of-the-art pretrained language models—BART, DialoGPT, and BERT. The proposed method is noted for its scalability and flexibility, enabling it to address hate speech across multiple languages and contexts.

\section{Impact of artificial intelligence on textual cyber abuse detection}
\label{sec:other}

After reviewing the most prominent papers on LLMs for cyberbullying or hate speech detection, we extended our analysis to explore the impact of these models, as well as traditional machine learning and deep learning techniques, on other types of cyber abuse. In this part of the survey, we broadened our scope beyond LLMs to include other artificial intelligence techniques, given that forms of cyber abuse like doxing and shaming are less frequently studied. This broader approach allowed us to identify a wider range of papers and derive more comprehensive insights. Despite this more general analysis, we remained focused on the same key questions as in the previous section, aiming to understand the influence of models like BERT in the context of textual cyber abuse. Table \ref{tab:overview} summarizes these findings. From the table, it is clear that transformer-based techniques are prominent, alongside other neural models like biLSTM, which, in this context, functions effectively as a language model due to its ability to handle sequence analysis in texts. Namely BERT/Transformer models are used in 29.41\% of the papers and LSTM/BiLSTM models are used in 35.29\% of the papers. SVM is the most common traditional technique, used in 23.53\% of the papers.

\begin{table}[htb]
\centering
\small
\begin{adjustbox}{max width=\textwidth}
\begin{tabular}{llllp{3.5cm}}
\hline
\textbf{CITE (Author, Year)} & \textbf{cyber abuse} & \textbf{Advanced NN Model /} & \textbf{Traditional} & \textbf{Unbalanced Data} \\ 
& \textbf{Type} & \textbf{LLM Used?} & \textbf{Techniques Used?} & \textbf{Processed?} \\ \hline
\citep{aarthi2022deep}        & Shaming               & DRNN                                    & SVM                                   & Not Handled   \\ 
\citep{aarthi2023hatdo}       & Doxing                & biLSTM, biGRU                           & SVM                                   & Partially Handled   \\ 
\citep{miao2024deep}            & Doxing                & deBERTa, BiLSTM                         & None                                  & Not Handled         \\ 
\citep{li2022covert}              & Doxing                & DNN                                     & Regularization                        & Handled Effectively including regularization terms to the training loss \\ 
\citep{WOS:001026644900007}            & Emotional Abuse       &  LSTM, GMM                                    & None                           & Partially handled, taken into account in the experimentation design  \\ 
\citep{WOS:000792675500001}            & Emotional Abuse       & BERT                                    & LDA                                   & Not Handled         \\ 
 \citep{WOS:000801709000001}            & Emotional Abuse       & None                                    & Logistic Regression, RF, ANN          &  Not Handled   \\
 \citep{bakar2023text}          & Trolling              & None                                    & SVM                                   & Handled Effectively with class weighting  and RBF kernel in SVM \\ 
\citep{nepomuceno2023should} & Trolling            & None                                    & LDA, MLP                              & Not Handled         \\ 
 \citep{cranmer2024social}    & Trolling              & BERT                                    & None                                  & Not Handled         \\ 
 \citep{ezzeddine2023exposing} & Trolling            & LSTM                                    & None                                  & Handled Effectively by means of under-sampling \\
\citep{asif2024graph}           & Trolling              & GCN, GloVe                              & None                                  & Not Handled  \\ 
\citep{swed2024keeping}         & Trolling              & None                                    & Random Forest                         & Handled Effectively compensating in the algorithm the unequal distribution  \\ 
 \citep{uyheng2022language}   & Trolling              & BERT, LSTM                              & Logistic Regression, SVM, RF          & Not Handled         \\ 
 \citep{macdermott2022using} & Trolling            & biLSTM, biGRU                           & None                                  & Handled Effectively by means of under-sampling \\ 
\citep{mewada2024cipf}        & Impersonation         & CNN                                     & None                                  &  Not Handled  \\ 
 \citep{huang2021stop}          & Impersonation         & None                                    & Gaussian Dist., Euclidean Dist.       & Not Handled         \\ \hline
\end{tabular}
\end{adjustbox}
\caption{Overview of other ways of textual cyber abuse Detection}
\label{tab:overview}
\end{table}

Finally, to gain a comprehensive understanding of the proposed papers, we conducted an analysis of their key findings, models, advancements, results, and the handling of unbalanced data. This analysis focused on the five subcategories of cyber abuse identified in our Web of Science queries.

\subsection{Shaming and Cancel Culture}

Regarding the detection of shaming and cancel culture in text using LMs or neural networks, only one comprehensive study was retrieved \citep{aarthi2022deep}. Aarthi and Balika presents a novel approach to detecting and classifying online shaming comments on social media, specifically Twitter, by utilizing a deep recurrent neural network (DRNN) optimized with the Aquila optimization algorithm \citep{abualigah2021aquila}, a population-based optimization technique inspired by the hunting behavior of aquila birds. The study employs a substantial dataset comprising approximately 1,600,034 tweets collected through various public APIs. The results demonstrate that the proposed DRNN model, enhanced by the Aquila optimization algorithm, significantly improves classification accuracy in identifying shaming comments compared to traditional methods. By focusing on the syntactic, semantic, and contextual features extracted from the tweets, the model effectively addresses the complexities of cyber abuse, such as sarcasm and context-dependent meanings. This research contributes to the field by providing a robust framework for detecting online shaming.

\subsection{Doxing}
Authors in \citep{aarthi2023hatdo} introduces a novel approach for detecting and classifying offensive comments on social media, with a specific focus on doxing. The model employs advanced techniques such as CNN for feature extraction and Bidirectional Gated Recurrent Units (BiGRU) for sequence modeling, allowing it to capture both local patterns and contextual information in the text. The study utilizes various datasets, including a labeled hate speech detection dataset with 3,000 comments from platforms like Twitter and Reddit, and the Tweepy dataset, which contains over a million tweets. The results demonstrate the model's strong performance, achieving an F1-score of 0.808 for English subtask A, indicating effective classification capabilities. Specifically, the model incorporates intrinsic methods to identify sensitive keywords and phrases associated with doxing, enhancing its ability to detect harmful content.

More recently, in \citep{miao2024deep} deep learning-based method for preventing data leakage in the Electric Power Industrial Internet of Things, specifically addressing the challenges of safeguarding sensitive personal data during interactions between the State Grid business platform and third-party platforms. The main contributions include the development of a robust framework that combines the DeBERTa \citep{he2020deberta} model with a BiLSTM-CRF \citep{huang2015bidirectional} architecture to enhance the identification of privacy-sensitive information, such as identity cards, phone numbers, and email addresses, within both structured and unstructured text data. The experimental results demonstrate an F1 score of 81.26\% on the CLUENER 2020 dataset \citep{xu2020cluener2020finegrainednamedentity}, indicating the method's effectiveness in accurately detecting sensitive data entities and mitigating the risk of data leakage. In terms of data handling, the approach employs a combination of regular expressions and keyword detection to construct a sensitive data distribution library, facilitating the identification and filtering of personal information. The model processes text data by extracting semantic features and contextual information, allowing for precise matching of sensitive content. This dual strategy not only enhances the recognition of personal data but also ensures that privacy-sensitive information is desensitized before being shared, thereby providing a reliable solution for data security in the power industry context.

Finally \citep{li2022covert} presents the covert task embedding (CTE) attack, a novel method that exploits deep neural networks to extract sensitive personal information while performing a legitimate task, such as age estimation. The main contribution lies in demonstrating how a deep neural networks can be trained to simultaneously execute a primary task and a covert task (e.g., detecting gender or ethnicity) without compromising the accuracy of the primary function. The experiments utilized face-based images as input data, showcasing the feasibility of the CTE attack across different nerual network architectures, including multiclass classification and soft-ranking networks. The results indicated that the hidden information could be reliably extracted without impairing the performance on the primary task, highlighting significant privacy risks. In terms of data handling, the study emphasizes the importance of safeguarding personal information by illustrating how covert tasks can be embedded within seemingly innocuous models. The authors propose a key-protected CTE approach, where the extraction of sensitive information requires a secret key, thereby adding a layer of complexity to the data leakage process. This method underscores the need for robust privacy measures in neural networks applications in contexts where personal data is processed.

\subsection{Emotional and Psychological Abuse}

Although the subsequent papers do not specifically target cyber abuse, they are relevant to our study due to their focus on text-based data and advanced analytical techniques. Although these studies do not directly address cyber abuse, their methodologies and findings provide a foundation that can be leveraged to enhance the detection and understanding of emotional abuse in digital environments. Thus, while not directly applicable, these papers contribute to the broader context and future development of cyber abuse detection tools.

The study \citep{WOS:000792675500001} investigates the use of NLP techniques to analyze text data for predicting personality traits and psychological distress. It combines qualitative data from written respons es and transcribed speech, employing models such as Latent Dirichlet Allocation (LDA) and BERT \citep{Devlin2018} for data processing and analysis. Although the study does not directly address cyber harassment, its methodologies could be adapted to study emotional abuse in online contexts by analyzing language patterns indicative of distress or victimization. The research highlights the importance of developing accurate algorithms for predicting psychological outcomes from various linguistic inputs, but it does not address the issue of imbalanced data. The findings could be used to inform the development of tools for detecting emotional abuse by identifying specific linguistic markers associated with distress in online communications.

The paper \citep{WOS:000801709000001} explores psychological distress by applying four machine learning algorithms—logistic regression, random forests, artificial neural networks, and gradient boosting—to data from the Health Information National Trends Survey (HINTS), which includes 5,484 respondents. The study identifies 20 significant predictors of psychological distress, including sociodemographic and lifestyle-related factors. The dataset was a bit unbalance, with 2,874 respondents reporting psychological distress and 2,610 without, but this scenario was correctly addressed using techniques such as 10-fold cross-validation to improve model robustness, and they provided metrics such as AUC and F1 score to evaluate model performance.

Although not focused solely on text, a study by \citep{WOS:001026644900007} examines speech features related to emotional distress. The research aims to develop an automated speech analysis algorithm capable of detecting clinically relevant emotional distress and functional impairments in children and adolescents with mental health disorders. The study's key contribution lies in its integration of speech analysis with validated psychological assessments, providing a comprehensive understanding of emotional states. Preliminary results are promising, with the algorithm achieving up to 85.35\% accuracy in emotion classification using a Gaussian mixture model (GMM). The GMM effectively handles unbalanced data by emphasizing feature extraction and frame selection techniques. While the study primarily focuses on detecting emotional distress, it also has implications for identifying issues related to cyber harassment and emotional abuse by recognizing emotional signals in speech that may originate from such experiences.

\subsection{Trolling}

Bakar et al. in \citep{bakar2023text} introduce a Support Vector Machine (SVM) model classification model aimed at detecting troll threat sentences in the Malay language, contributing significantly to the field of text simplification and addressing the challenges of cyber abuse associated with trolling. The study utilized the Malay Text Simplification Dataset, comprising 6,836 instances categorized as complex or non-complex, and implemented class weighting in the SVM classifier to effectively tackle the issue of class imbalance within the dataset. This approach ensures that both complex and non-complex sentences are accurately identified, enhancing the model's performance. Results indicated that the SVM classifier achieved an impressive average accuracy of 92.37\% when incorporating semantic features. The model's focus on semantic and lexical features allows for a nuanced understanding of trolling behavior, which often involves complex language and hidden meanings.

In their study, Marcelo Vinhal et al. \citep{nepomuceno2023should} examine the impact of marketer-generated content (MGC)—marketing communications initiated by companies on their official social media pages—on user interactions in online communities, with a particular focus on trolling. The authors analyze a large dataset of posts from platforms such as Facebook, Instagram, and Twitter, employing a mix of unsupervised and supervised machine learning techniques, including Latent Dirichlet Allocation (LDA) \citep{blei2003latent} and vector autoregression (VAR) to classify content and assess its toxicity. Their findings reveal a positive correlation between the average toxicity of comments and user engagement, suggesting that toxic interactions may inadvertently boost product usage. To address the challenge of toxicity classification, the study employs a time-series model (panel-data vector autoregression) to effectively capture the dynamics of user interactions over time. However, despite the unbalanced nature of the problem and the dataset, the study does not provide evaluations related to this imbalance. In the context of cyber abuse, the research underscores the psychological motivations behind trolling behavior, which often involve traits like narcissism and sadism. Trolling is identified as the intentional disruption of online communities through toxic comments. The authors stress the importance of understanding the elements of MGC that may contribute to such negative interactions and advocate for strategies to mitigate trolling behavior.

In \citep{cranmer2024social}, the authors contribute to the understanding of state-sponsored disinformation and the role of social media trolls in shaping public discourse. The study analyzes a dataset of 163,549 tweets related to Daryl Morey’s controversial tweet supporting pro-democracy protests in Hong Kong. The research categorizes these tweets into various themes, revealing that 68.81\% of the responses were focused on promoting the image of the People's Republic of China (PRC) and attacking Morey. This highlights the significant prevalence of troll activity and its impact on public perception. To address the challenge of unbalanced data, the authors employed a natural language classifier, achieving an overall accuracy of 78\% in categorizing tweets, comparable to the performance of the BERTweet model \citep{nguyen2020bertweet}. The dataset was randomly sampled and categorized into five distinct themes, some of which had low representation. However, the paper does not provide detailed explanations regarding the handling of data imbalance. The study emphasizes the need for advanced analytical frameworks to effectively combat disinformation and online abuse. By leveraging machine learning and NLP techniques, the authors successfully identified and categorized troll activity, offering valuable insights into how these tactics distort online discourse.

In line with precedent work examining how trolls are utilized by foreign entities, \citep{swed2024keeping} advances our understanding of information operations by highlighting the distinct social footprint of Russian troll accounts, shaped by specific social norms and expectations. The study leverages extensive datasets, including over 2.9 million English tweets and nearly 4.9 million Russian tweets from the Internet Research Agency (IRA), focusing on text data to analyze behavioral patterns. The research categorizes accounts into four types: Fake News, Organizations, Political Affiliates, and Default Individuals. The model is built on a random forest classifier (RFC), utilizing  carefully selected features to minimize multicollinearity, thereby improving the model’s learning efficiency. To address the challenge of imbalanced data, the RFC algorithm compensates for the unequal distribution of categories within the dataset. Additionally, least absolute deviation (L1) normalization is applied to balance the weights of each variable, preserving the dataset's structure while enhancing the model's predictive accuracy. The model achieved high accuracy rates, up to 90.7\% for certain categories, by leveraging features such as tweet frequency, timing, and linguistic characteristics to identify and predict the behavior of troll accounts. This approach is particularly relevant to understanding cyber abuse, as it provides a robust framework for detecting and analyzing trolling behaviors that frequently manifest as targeted harassment on social media platforms.

Ezzeddine et al. in \citep{ezzeddine2023exposing} introduce a novel behavioral-based approach for detecting state-sponsored troll activity on social media, emphasizing the analysis of online activity sequences rather than just textual content. The key contributions of the study include the development of a ``Troll Score'', a metric designed to quantify troll-like behavior, and the implementation of an LSTM model for classifying user trajectories with high accuracy. The study utilizes a dataset from Twitter, examining both active and passive online activities to distinguish the behavioral patterns of trolls from those of organic users. These sequences, termed ``trajectories'', capture the chronological order of actions such as posting, retweeting, and interacting with others. The results highlight the model's effectiveness, achieving an AUC of approximately 91\% in differentiating trolls from legitimate users, even in the presence of unbalanced data—demonstrating the robustness of the model’s performance. The study addresses the challenges of cyber abuse related to trolling by focusing on behavioral cues that are more difficult to mimic than textual content, thereby increasing the resilience of their detection method against the sophisticated tactics employed by LLMs in influence operations.

Asif et al. in \citep{asif2024graph} introduce a novel machine learning framework designed to detect trolls and toxic content on social media by leveraging deep learning techniques, specifically Graph Convolutional Networks (GCNs) and GloVe word embeddings. A key innovation of this work is its ability to analyze text embedded within images, effectively addressing a significant limitation in existing content moderation methods. The study employs diverse datasets encompassing both textual and visual elements, acknowledging the challenges posed by imbalanced data, where non-toxic comments vastly outnumber toxic ones. To counter this imbalance, the authors apply data augmentation techniques, which significantly enhance the model’s ability to identify trolling behaviors. The results demonstrate that the GCN-based model achieves a testing accuracy of 0.92 and an F1-score of approximately 0.4, indicating its effectiveness in detecting toxic content, though with noted areas for improvement in precision and recall. The framework underscores the importance of contextual analysis in text data, utilizing natural language processing techniques to capture the subtleties of online interactions.

In \citep{uyheng2022language}, the authors advance our understanding of online trolling by employing a psycholinguistic framework to differentiate between trolls and bots, emphasizing that not all trolls are automated accounts and vice versa. The study uses a dataset of 4,917 labeled instances of trolling and non-trolling interactions, focusing on psycholinguistic features such as toxicity, cognitive complexity, and direct targeting. Leveraging these features, the researchers propose a classification system that utilizes both traditional machine learning models and advanced deep learning techniques, including BERT. The study provides insights into the nature of cyber abuse by analyzing the linguistic properties of trolling behavior. By examining the language used in trolling, the research lays the groundwork for developing tools to effectively identify and mitigate instances of cyber abuse. The authors found that a LSTM model achieved the best results. However, the classes in the dataset were manually balanced, which limits the generalizability of the system. Despite this limitation, the study underscores the importance of understanding the underlying psycholinguistic characteristics of malicious online interactions to create more effective detection and prevention strategies.

Finally in \citep{macdermott2022using}, the authors present a framework for detecting trolls and toxic content on social media by integrating deep learning techniques for both text and image analysis. The framework utilizes GloVe word embeddings and recurrent neural networks (RNNs), including Bidirectional LSTM and GRU, to analyze textual content. It employs a dataset of 223,549 comments from the Jigsaw dataset, which is characterized by a significant class imbalance, with less than 7\% of comments labeled as toxic. To address this imbalance, the authors apply undersampling techniques. The results indicate that the proposed models achieved high detection rates for toxic comments. However, accuracy decreased when analyzing text extracted from images due to the challenges associated with this task. The framework’s ability to extract and classify text from images is particularly important for addressing cyber abuse, as it facilitates the analysis of content in non-traditional formats such as memes or screenshots. The highest F-measure achieved was 0.88, using a ensemble model, supported by undersampling techniques.

\subsection{Impersonation}

In the context of impersonation, Mewada et al. \citep{mewada2024cipf} address the issue from a dual perspective: analyzing both the content generated by accounts and their interactions within networks. The study leverages techniques from text analysis and network analysis, particularly focusing on homophily—the tendency of individuals to associate with others who are similar in interests, demographics, or behaviors. This approach is used to propose a novel method for identifying fake profiles on social media, specifically targeting cyber abuse through impersonation and other malicious activities. The primary contribution of the paper is the integration of linguistic features, profile-centric features, and group-centric features to improve the detection of sockpuppets and crowdturfing communities. The study utilizes a dataset of user-generated content from various social media platforms, employing a Convolutional Neural Network (CNN) to classify users as either malicious or genuine. The results demonstrate the framework's effectiveness, with the CIPF framework achieving high precision, recall, and F1-scores, significantly outperforming state-of-the-art techniques in detecting fake profiles. A key aspect of the framework is its incorporation of linguistic features, including sentiment analysis, grammatical quality, and emotional context, which provide insights into the language patterns used by individuals. These linguistic features are critical for detecting impersonation, as they help identify anomalies in user-generated content that may indicate deceptive practices. This enhances the framework's ability to distinguish between genuine users and malicious actors, making it a robust tool for combating cyber abuse through impersonation.

We identified a borderline paper by Huang et al. \citep{huang2021stop}, where the authors proposed an automatic system to detect deceptive voices used in impersonation attempts. The reason for selecting this study is its two-fold approach, which analyzes both text-independent and text-dependent features. The latter involves detecting when an impersonator mimics specific phrases or commands typically used by the genuine speaker, thus linking the process to text analysis. The paper introduces a quasi-Gaussian distribution (QGD) model that effectively addresses both types of impersonation attacks on smart devices. The authors collected a unique dataset from a reality TV show, which enhances the robustness of their findings and enables a comprehensive evaluation of voice and text features. The results demonstrate that the QGD model can accurately distinguish between genuine and impersonated voices by analyzing the Euclidean distance within the voice (and text) feature space. The study reveals that voice features from the same individual are tightly clustered, while those from different individuals are more dispersed, making it a promising approach for detecting impersonation.

\section{Discussion}
\label{sec:discussion}

In this section, we discuss the results of our survey, focusing specifically on the research questions guiding this study. To enhance clarity and the practical utility of our findings, we have structured the discussion by addressing each research question individually. This approach ensures a comprehensive and organized analysis of the key insights drawn from the reviewed literature.

\subsection{RQ1: How representative are LMs and LLMS in detecting textual cyber abuse compared to traditional methods?}

To address this question, we performed an occurrence analysis based on the models used across the reviewed literature, focusing on those referenced in at least three papers. Table \ref{tab:lms_occurrences_filtered} presents the results for Language Models (LMs) and Large Language Models (LLMs), while Table \ref{tab:machine_occurrences_filtered} shows the results for traditional machine learning techniques. As seen in the tables, BERT emerges as the most widely utilized model, significantly surpassing other models, whether traditional machine learning techniques or LMs. Among LLMs, GPT stands out as the only model with multiple occurrences, whereas models like T5 and LLAMA appear only once.

When comparing the adoption of LMs to LLMs, we can observe that LMs remain the dominant tool in cyber abuse detection. BERT-based models are particularly widespread due to their flexibility in training and fine-tuning, making them highly effective. While LLMs are underrepresented, they present promising opportunities, especially for few-shot learning and data augmentation. Traditional machine learning techniques, though less prevalent in hate speech or cyberbullying detection, continue to play a role in other forms of cyber abuse. Notably, SVM and random forests remain useful due to their interpretability and ability to explain results. Interestingly, even older embedding techniques like FastText and Word2Vec still appear frequently in recent literature, highlighting their enduring relevance in specific contexts.

\begin{table}[htb]
\centering
\small
\begin{adjustbox}{max width=\textwidth}
\begin{tabular}{l l p{10cm}}

\hline
\textbf{Model} & \textbf{Occurrences} & \textbf{Citations} \\
\hline
BERT & 34 & \citep{muneer2023cyberbullying}, \citep{paul2023covid}, \citep{mahmud2024exhaustive}, \citep{li2024integrating}, \citep{alfurayj2024chained}, \citep{al2024towards}, \citep{maity2022multitask}, \citep{mahdi2024enhancing}, \citep{samee2023safeguarding}, \citep{saini2023enhancing}, \citep{ahmed2022performance}, \citep{ansari2024data}, \citep{zhou2023automated}, \citep{garcia2023evaluating}, \citep{garcia2023leveraging}, \citep{del2023socialhaterbert}, \citep{pookpanich2024offensive}, \citep{arcila2022detect}, \citep{maity2024hatethaisent}, \citep{anjum2024hatedetector}, \citep{nandi2024combining}, \citep{putra2024semi}, \citep{yun2023bert}, \citep{chhikaraidentification}, \citep{saleh2023detection}, \citep{almaliki2023abmm}, \citep{ramos2024leveraging}, \citep{pan2024comparing}, \citep{bilal2023roman}, \citep{ali2022hate}, \citep{firmino2024improving}, \citep{toliyat2022asian}, \citep{khan2022bichat}, \citep{althobaiti2022bert}, \citep{gupta2021bert} \\

bi-LSTM & 13 & \citep{muneer2023cyberbullying}, \citep{chen2024chinese}, \citep{paul2023covid}, \citep{mahmud2024exhaustive}, \citep{alfurayj2024chained}, \citep{al2024towards}, \citep{maity2022multitask}, \citep{ansari2024data}, \citep{maity2024hatethaisent}, \citep{saleh2023detection}, \citep{bilal2023roman}, \citep{mazari2024bert}, \citep{miao2024deep} \\

RoBERTa & 9 & \citep{alrowais2024robertanet}, \citep{ahmed2022performance}, \citep{su2023ssl}, \citep{yun2023bert}, \citep{pan2024comparing}, \citep{arshad2023uhated}, \citep{ali2022hate}, \citep{gupta2021bert}, \citep{pookpanich2024offensive} \\

DistilBERT & 8 & \citep{ahmed2022performance}, \citep{pookpanich2024offensive}, \citep{MontesinosCnovas2023SpanishHD}, \citep{sreelakshmi2024detection}, \citep{ali2022hate}, \citep{pan2024comparing}, \citep{malik2024hate}, \citep{firmino2024improving} \\

m-BERT & 5 & \citep{razi2024multilingual}, \citep{garcia2023leveraging}, \citep{MontesinosCnovas2023SpanishHD}, \citep{karayiugit2022homophobic}, \citep{ollagnier2023unsupervised} \\

BETO & 5 & \citep{garcia2023evaluating}, \citep{del2023socialhaterbert}, \citep{garcia2023leveraging}, \citep{MontesinosCnovas2023SpanishHD}, \citep{pookpanich2024offensive} \\

XLM-RoBERTa & 4 & \citep{mahmud2024exhaustive}, \citep{pookpanich2024offensive}, \citep{pan2024comparing}, \citep{ali2022hate} \\

GRU & 4 & \citep{mahmud2024exhaustive}, \citep{keya2023g}, \citep{aarthi2023hatdo}, \citep{mazari2024bert} \\

TwHIN-BERT & 3 & \citep{garcia2023leveraging}, \citep{pookpanich2024offensive}, \citep{MontesinosCnovas2023SpanishHD} \\

BiGRU & 3 & \citep{maity2022multitask}, \citep{wullach2022character}, \citep{mazari2024bert} \\
GPT (including GPT-2, GPT-3.5) & 3 & \citep{ahmed2022performance}, \citep{chhikaraidentification}, \citep{ramos2024leveraging} \\
\hline

\end{tabular}
\end{adjustbox}
\caption{Most used models LMs and LLMs models with more than 3 occurrences}
\label{tab:lms_occurrences_filtered}
\end{table}

Analyzing the performance of hate speech detection and other cyber abuse detection techniques presents challenges due to the lack of a standardized benchmark across studies. Many papers introduce new techniques and validate them on custom datasets or adapt them for specific languages, making direct comparisons difficult. This fragmentation in datasets and evaluation methods complicates the ability to gauge overall progress in the field.

However, from a broader perspective, the use of transformer-based architectures consistently outperforms other approaches. This trend is evident in widely used datasets like Ethos Binary \citep{mollas2020ethos}, HateXplain \citep{mathew2021hatexplain}, and OffensEval 2019 \citep{zampieri2019semeval}, where BERT-based models consistently top the leaderboards \citep{rajput2021hate, caselli-etal-2021-hatebert, kim-etal-2022-hate}, reaffirming their superior performance in cyber abuse detection tasks.

\begin{table}[htb]
\centering
\small
\begin{adjustbox}{max width=\textwidth}
\begin{tabular}{l l p{10cm}}

\hline
\textbf{Machine learning or traditional embeddings} & \textbf{Occurrences} & \textbf{Citations} \\
\hline
SVM & 10 & \citep{paul2023covid}, \citep{mahmud2024exhaustive}, \citep{ahmed2022performance}, \citep{saini2023enhancing}, \citep{zhou2023automated}, \citep{arcila2022detect}, \citep{sreelakshmi2024detection}, \citep{althobaiti2022bert}, \citep{aarthi2022deep}, \citep{aarthi2023hatdo} \\

Logistic Regression (LR) & 9 & \citep{paul2023covid}, \citep{mahmud2024exhaustive}, \citep{li2024integrating}, \citep{mehta2022social}, \citep{arcila2022detect}, \citep{sreelakshmi2024detection}, \citep{anjum2024hatedetector}, \citep{bilal2023roman}, \citep{uyheng2022language} \\

Random Forest (RF) & 8 & \citep{mahmud2024exhaustive}, \citep{li2024integrating}, \citep{sreelakshmi2024detection}, \citep{mehta2022social}, \citep{bilal2023roman}, \citep{toliyat2022asian}, \citep{husain2022investigating}, \citep{swed2024keeping} \\

CNN & 7 & \citep{ansari2024data}, \citep{putra2024semi}, \citep{bilal2023roman}, \citep{khan2022bichat}, \citep{mewada2024cipf}, \citep{paul2023covid}, \citep{saini2023enhancing} \\

Naive Bayes (NB) & 5 & \citep{mahmud2024exhaustive}, \citep{sreelakshmi2024detection}, \citep{arcila2022detect}, \citep{mehta2022social}, \citep{toliyat2022asian} \\

XGBoost & 5 & \citep{ahmed2022performance}, \citep{yun2023bert}, \citep{bilal2023roman}, \citep{toliyat2022asian}, \citep{li2024integrating} \\

Word2Vec & 3 & \citep{muneer2023cyberbullying}, \citep{saleh2023detection}, \citep{altinel2024so} \\

FastText & 3 & \citep{paul2023covid}, \citep{maity2024hatethaisent}, \citep{arshad2023uhated} \\
\hline

\end{tabular}
\end{adjustbox}
\caption{Most used traditional machine learning methods with more than 3 occurences}
\label{tab:machine_occurrences_filtered}
\end{table}

\subsection{RQ2: Considering that cyber abuse frequently involves imbalanced classification challenges, are researchers adequately addressing this issue in their evaluation metrics and pre-processing methods?}

Based on the reviewed literature, it is clear that researchers in the field of cyber abuse detection are increasingly recognizing the challenges posed by imbalanced classification. Many studies now utilize a range of evaluation metrics—such as precision, recall, and F1-score—which provide a more nuanced understanding of model performance than accuracy alone, particularly in contexts where class distribution is skewed. However, some research still relies on overall accuracy or non-weighted metrics, potentially obscuring the true effectiveness of models in detecting minority classes.

A notable trend is the adoption of various pre-processing methods to address class imbalance, including oversampling techniques like SMOTE \citep{li2024integrating} and undersampling methods. These strategies aim to create a more balanced training dataset, ultimately improving model performance on underrepresented classes. Additionally, some studies are incorporating cost-sensitive learning, which assigns varying penalties for misclassifications based on the importance of each class. The introduction of generative models, such as GPT, for data augmentation also represents a promising avenue for enhancing the minority class representation, functioning similarly to oversampling techniques \citep{wullach2022character}.

Despite these advancements, a lack of standardization in how imbalanced classification is addressed remains a significant issue across studies. This inconsistency complicates efforts to compare results and identify best practices. While there is growing awareness of the need to tackle class imbalance in cyber abuse detection, substantial gaps persist. Improved standardization of evaluation metrics and preprocessing methods, coupled with clearer reporting of methodologies, would greatly enhance the robustness and comparability of findings in future research.

\subsection{RQ3: Based on the reviewed literature, what emerging trends and future directions are shaping the development of cyber abuse detection?}

One key insight derived from the survey is that recent advancements in cyber abuse detection are focused more on applying standardized English-language models to new or underrepresented languages, rather than developing entirely new architectures. The predominant approach in these studies involves transfer learning and fine-tuning existing models, which allows for the adaptation of well-established models to different linguistic contexts. This trend reflects a shift towards leveraging the strengths of existing architectures, optimizing them for diverse languages rather than creating novel frameworks from scratch.

In cyber abuse detection, several evolving trends are reshaping the field. Traditionally, this domain has been treated primarily as a binary classification problem—identifying abusive or non-abusive content. However, more recent studies are shifting towards a more nuanced understanding, recognizing that cyber abuse involves multiple dimensions. These include factors such as the nature of the target, the directness of hate, the type of hostility, and whether the abuse is community-driven. This shift has led to the development of multi-task algorithms, capable of performing tasks such as sentiment analysis, community detection, and toxicity prediction simultaneously.

A growing trend is the incorporation of social network analysis and traditional data mining techniques such as sentiment analysis. These methods are increasingly acknowledged as powerful tools for analyzing the complex, relational aspects of cyber abuse, such as how it spreads within communities or networks and the influence of key actors in propagating hate speech. We identified four insightful papers that highlight the value of unsupervised techniques in this area. These works underscore the necessity of analyzing cyber abuse from multiple perspectives, beyond simple classification, to capture its broader dynamics and impact.  

If we stick only to classification domain of the problem, the majority of efforts are focused on text preprocessing and the creation of embeddings to capture the nuances of language. Among various models, the combination of BERT and LSTMs has emerged as one of the most powerful tools for this purpose.

Another significant observation is that the majority of current datasets are heavily based on Twitter, while other social platforms, such as Reddit, remain underexplored. This reliance on a single platform limits the generalizability and versatility of existing models. Expanding datasets to include diverse sources of social media data is crucial for enhancing the robustness and adaptability of cyber abuse detection systems.

Finally, there is a growing call for the standardization of evaluation processes across studies. Many papers claim to outperform baseline models, but these claims often rely on custom datasets or datasets limited to a single platform, making meaningful comparisons difficult. A standardized approach to reporting preprocessing methods, evaluation metrics, and dataset usage is essential. This would improve the comparability of findings across different studies and support the development of best practices in the field, ensuring more reliable and generalizable models for detecting cyber abuse.

\subsection{RQ4: Which types of cyber abuse detection are most significantly influenced by LLMs?}

As demonstrated in Table \ref{tab:lms_occurrences_filtered}, BERT-based techniques continue to dominate across various types of cyber abuse detection. However, when focusing solely on LLMs, certain areas—such as shaming, doxing, emotional abuse, trolling, and impersonation—remain largely unexplored. In the domain of cyberbullying, only two papers \citep{ferrer2024using,ahmed2022performance} have utilized LLMs, while in hate speech detection, six papers \citep{wullach2022character,pan2024comparing,ramos2024leveraging,chhikaraidentification,garcia2023leveraging,ahmed2022performance} are based on LLMs. 

These findings lead us to two conclusions. First, hate speech is the form of cyber abuse most influenced by LLMs. This may, however, be partially attributed to the huge focus on hate speech within the research community, making it the most extensively studied area in cyber abuse detection. Second, the application of LLMs in cyber abuse detection is still in its infancy and warrants further exploration, especially given the potential of LLMs demonstrated in other fields.
 
\subsection{RQ5: Are there specific forms of cyber abuse that remain underexplored and lack adequate detection solutions?}

To conclude this section, it is crucial to recognize the significant gaps in research concerning other kinds of cyber abuse as emotional abuse and doxxing within the broader landscape of textual cyber abuse detection and analysis. Despite the serious harm that these forms of cyber abuse can inflict, they remain underexplored and are often overshadowed by a predominant focus on hate speech. While comprehensive studies on hate speech may touch upon related forms of abuse, there is a pressing need for more specialized and nuanced approaches tailored to address these specific issues.

The situation is further exacerbated in other forms of cyberbullying, such as dogpiling—an online harassment tactic where groups of individuals collectively target the same victim. This type of cyberbullying could benefit significantly from social network analysis and data mining techniques, yet it remains largely unexamined in the literature.

Similarly, in the domains of doxing and emotional or psychological abuse, the specific targets and contexts have received insufficient attention. Most existing research relies heavily on classification techniques; however, traditional unsupervised methods could be very valuable for analyzing psychological \citep{rosenbusch2021supervised} and emotional abuse. Techniques such as similarity detection and correlation analysis could also enhance our understanding of doxing and its implications \citep{alonso2021writer}.

\section{Conclusion}
\label{sec:conclusion}

In this paper, we have reviewed the latest research on textual cyber abuse detection, with a particular focus on studies that leverage LMs and LLMs. This survey is the first to extend its scope beyond commonly studied areas such as hate speech and cyberbullying, encompassing a broader range of cyber abuse typologies.

We analyzed 68 papers, rigorously categorizing them according to different forms of cyber abuse and their harmful objectives. From this body of work, we identified critical insights into novel methodologies and the increasing role of language models, which have become essential in contemporary cyber abuse detection.

Our findings emphasize the widespread dominance of BERT-based approaches, particularly in the detection of hate speech, where these models consistently achieve state-of-the-art results. BERT has emerged as the most extensively used model for text extraction and classification in the field.

Additionally, emerging LLMs like GPT and FLAN-T5 show significant promise, especially in low-resource settings and few-shot learning scenarios, where they excel at capturing unseen linguistic nuances. Generative models like GPT also hold potential for data augmentation, offering a dual benefit: the creation of synthetic datasets for training and mitigating issues of class imbalance.


\section{Acknowledgments}
\label{Acknowledgments}

The research reported in this paper was supported by the DesinfoScan project: Grant TED2021-129402B-C21 funded by MICIU/AEI/10.13039/ 501100011033 and by the European Union NextGenerationEU/PRTR, and FederaMed project: Grant PID2021-123960OB-I00 funded by MICIU/ AEI/ 10.13039/ 501100011033 and by ERDF/EU. Finally, the research reported in this paper is also funded by the European Union (BAG-INTEL project, grant agreement no. 101121309).

\bibliographystyle{apalike}
\bibliography{mybibfile}

\end{document}